\documentclass{article}

 \usepackage[preprint]{neurips_2026}

\usepackage[utf8]{inputenc} %
\usepackage[T1]{fontenc}    %
\usepackage{hyperref}       %
\usepackage{url}            %
\usepackage{booktabs}       %
\usepackage{amsfonts}       %
\usepackage{nicefrac}       %
\usepackage{microtype}      %
\usepackage{xcolor}         %
\usepackage{enumitem}

\usepackage{algorithm}
\usepackage{algorithmic}

\usepackage{amsmath}
\usepackage{amssymb}
\usepackage{mathtools}
\usepackage{amsthm}
\usepackage{subcaption}

\definecolor{C0}{HTML}{66C2A5}
\definecolor{C1}{HTML}{FC8D62}
\definecolor{C2}{HTML}{8DA0CB}
\definecolor{CommentGreen}{HTML}{0D730D}

\newcommand{\comm}[1]{\textcolor{gray}{{\# #1}}}
\newcommand{\com}[1]{\hfill \textcolor{gray}{{\# #1}}}

\renewcommand{\paragraph}[1]{\textbf{#1}}

\newcommand{\methodlong}{Automated Background Swapping}
\newcommand{\method}{\texttt{AutoBackSwap}}

\usepackage{color}
\usepackage{amsmath}
\usepackage{amssymb}
\usepackage{amsthm}
\usepackage{mathtools}
\usepackage{thmtools}
\usepackage[mathscr]{eucal}
\usepackage{bm}
\usepackage{bbm}
\usepackage{relsize}
\usepackage{graphics}
\usepackage{wrapfig}
\usepackage{multirow}
\usepackage{enumitem} 
\usepackage{cancel}
\usepackage{todonotes}
\usepackage[algo2e]{algorithm2e}

\usepackage[capitalize,noabbrev]{cleveref}

\usepackage{array}

\usepackage{pgf}
\usepackage{xinttools}
\usepackage{tikz}
\usetikzlibrary{arrows.meta}
\usepackage{textcomp}
\usetikzlibrary{calc,shapes,arrows,positioning,automata,trees}
\usepackage{placeins} 
\usepackage{tabularx}
\usepackage{graphicx}
\usepackage{subcaption} 
\usepackage{listings}
\usepackage{xargs,lipsum,caption,changepage,ifthen}

\usepackage{tikz} 

\usepackage{footmisc}

\usepackage{bigdelim}
\usepackage{colortbl} 
\definecolor{mColor1}{rgb}{0.95,0.95,0.95}
\newcolumntype{a}{>{\columncolor{mColor1}}c}

\usepackage[toc,page]{appendix}

\usepackage{tcolorbox}

\definecolor{solarized@base03}{HTML}{002B36}
\definecolor{solarized@base02}{HTML}{073642}
\definecolor{solarized@base01}{HTML}{586e75}
\definecolor{solarized@base00}{HTML}{657b83}
\definecolor{solarized@base0}{HTML}{839496}
\definecolor{solarized@base1}{HTML}{93a1a1}
\definecolor{solarized@base2}{HTML}{EEE8D5}
\definecolor{solarized@base3}{HTML}{FDF6E3}
\definecolor{solarized@yellow}{HTML}{B58900}
\definecolor{solarized@orange}{HTML}{CB4B16}
\definecolor{solarized@red}{HTML}{DC322F}
\definecolor{solarized@magenta}{HTML}{D33682}
\definecolor{solarized@blue}{HTML}{268BD2}
\definecolor{solarized@cyan}{HTML}{2AA198}
\definecolor{solarized@green}{HTML}{859900}

\newtcolorbox{importantresult}{colback=solarized@yellow!5!white,
colframe=solarized@yellow,parbox, left=0.5mm, right=0.5mm,top=0.5mm,bottom=0.5mm}

\newtcolorbox{importantresult_noparbox}{colback=solarized@yellow!5!white,
colframe=solarized@yellow,parbox=false, left=0.5mm, right=0.5mm,top=0.5mm,bottom=0.5mm}

\newcommand\Bm{\bm{m}}

\newcommand\Bx{\bm{x}}

\newcommand\Bz{\bm{z}}

 \newcommand{\dR}{\mathbb{R}}

\newcommand{\cA}{\mathcal{A}} \newcommand{\cB}{\mathcal{B}}
 \newcommand{\cD}{\mathcal{D}}
 \newcommand{\cF}{\mathcal{F}}

\newcommand{\cM}{\mathcal{M}}

 \newcommand{\cX}{\mathcal{X}}
\newcommand{\cY}{\mathcal{Y}}

\DeclarePairedDelimiterX{\kldiv}[2]{(}{)}{%
  #1\;\delimsize\|\;#2%
}

\DeclarePairedDelimiterX{\mi}[2]{(}{)}{%
  #1\;\delimsize ; \;#2%
}

\DeclarePairedDelimiterX{\di}[2]{(}{)}{%
  #1\;\delimsize ; \;#2%
}

\DeclarePairedDelimiterX{\ce}[2]{(}{)}{%
  #1\;\delimsize ; \;#2%
}

\DeclarePairedDelimiterXPP{\mii}[3]%
   {_{\mathrm{#1}}}{(}{)}{}{#2\;\delimsize ; \;#3%
}

\makeatletter
\newcommand{\dlmf}[1]{%
\citep[%
  \def\nextitem{\def\nextitem{, }}%
  \@for \el:=#1\do{\nextitem\href{http://dlmf.nist.gov/\el}{(\el)}}%
]{Olver:10}%
}
\makeatother

\usepackage{pdflscape} 

\newcolumntype{R}[1]{>{\raggedright\arraybackslash}p{#1}}
\newcolumntype{C}[1]{>{\centering\arraybackslash}p{#1}}
\newcolumntype{L}[1]{>{\raggedleft\arraybackslash}p{#1}}

\usepackage{bigdelim}
\usepackage{colortbl} 
\definecolor{mColor1}{rgb}{0.95,0.95,0.95}

\hypersetup{
   colorlinks=true,
   linkcolor=[RGB]{80, 100, 180}, %
   citecolor=[RGB]{80, 100, 180},
   urlcolor=magenta,
   pdfborder=0 0 0,
   pdftitle={},
   pdfsubject={}, 
   pdfkeywords={},
   pdfauthor={},%
   pdfstartview=FitH
}

\usepackage{arydshln}

\title{Automated Background Swapping for Robustness against Spurious Backgrounds}

\author{%
  Cesar Roder \\
  Johannes Kepler University Linz, Austria\\
  \texttt{cesar\_roder@outlook.at} \\
  \And
  Kajetan Schweighofer \\
  Cognizant AI Lab, USA \thanks{Work partly done while at Johannes Kepler University Linz, Austria} \\
  \texttt{kai.schweighofer@gmx.at} \\
}

\begin{document}

\maketitle

\begin{abstract}
  Classifiers based on Deep Neural Networks exhibit strong performance across domains, yet can fail catastrophically if they rely on spurious correlations, i.e., features that are predictive of the target label in the training data but are not causally linked and thus fail to generalize.
  For the vision domain, many such spurious correlations manifest themselves within the background of the image, where only the foreground is predictive of the class label.
  In this paper, we introduce \methodlong{} (\method{}) to reduce the reliance of classifiers on such spurious backgrounds.
  \method{} uses a secondary network to disentangle the foreground and background, followed by infilling to synthesize complete backgrounds, and finally combines different foregrounds and inpainted backgrounds to augment the training data.
  We find that patch-wise labeling of just a few hundred samples suffices to train the secondary network and automatically augment the full training dataset on challenging image classification tasks.
  In contrast to many previous methods, \method{} proves very effective even if there is not a single sample in the training data breaking the spurious correlation.
  Across a range of image classification tasks with spurious backgrounds, \method{} consistently outperforms prior methods.
\end{abstract}

\section{Introduction}

Deep neural networks achieve state-of-the-art performance on image classification benchmarks, yet they often succeed by exploiting spurious correlations: visual cues that correlate with labels but are not causally related.
When such correlations break at inference time, performance can degrade severely, with serious consequences in real-world applications \citep{Buolamwini:18, Oakden-Rayner:20}.
Spurious correlations often manifest themselves within the background of the image \citep{Sagawa:20, Xiao:21, Moayeri:22}, as backgrounds tend to have simple features and often dominate the overall image.
For example, classifiers distinguishing cows from camels may rely on grass vs. sand in the background scene rather than animal morphology, if background cues are easier to learn than semantic object features \citep{Scimeca:22}.

\begin{figure*}[htp!]
    \centering
    \includegraphics[width=0.92\linewidth]{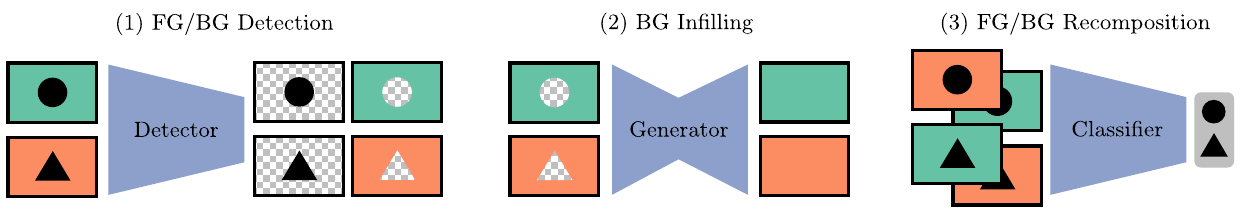}
    \caption{Three main steps of \method{}. In this example, $\cY = \{\bullet, \blacktriangle\}$ and $\cA = \{\mathcolor{C0}{\blacksquare}, \mathcolor{C1}{\blacksquare} \}$. The detector disentangles foreground and background by predicting a binary mask. Then, the generator inpaints the missing parts of the background to form a full background image. Finally, foreground and background are recombined stochastically to train a classifier that is invariant to the background.}
    \label{fig:method}
    \vspace{-0.2cm}
\end{figure*}

A broad range of prior work has been proposed to mitigate spurious correlations, including worst-group risk optimization \citep{Sagawa:20}, invariant representation learning \citep{Arjovsky:19}, and penalties that reduce sensitivity to spurious attributes \citep{Sun:16b, Ganin:16}.
Other approaches rely on retraining using auxiliary data, either based on the original network \citep{Liu:21, Zhang:22, Qiu:23} or from held-out reweighting datasets \citep{Kirichenko:23}.
Despite their differences, many of these methods implicitly assume access to samples in which the correlation between the target label and the spurious background is broken during training.
In practice, collecting such samples can be prohibitively expensive or infeasible, especially when certain combinations of target and spurious attributes are rare or physically difficult to obtain.

In this work, we propose \methodlong{} (\method{}), a data-augmentation-based approach that removes the need for observing broken correlations during training.
\method{} consists of three main steps: \textbf{(1)} disentangling foreground and background, \textbf{(2)} generating complete background images via infilling, and \textbf{(3)} stochastically recomposing foregrounds and backgrounds to train a classifier that is invariant to background cues (see Fig.~\ref{fig:method}).
The key insight is that very few samples are sufficient to train an auxiliary model to predict foreground vs. background with sufficient quality on the training data.
In our experiments, a few hundred per dataset suffice, e.g. 287 for Waterbirds \citep{Sagawa:20}; about 5\% of the training data.
Moreover, we show that off-the-shelf generative models are more than adequate for infilling of background images, and even simplistic strategies such as pixel scrambling or constant infilling can be reasonable alternatives with minimal impact on the effectiveness of \method{}.

Our main contributions in this work are

\begin{itemize}[leftmargin=*, topsep=0pt, itemsep=4pt, partopsep=0pt, parsep=0pt]
    \item We propose \method{}, a scalable data augmentation approach that reduces the dependence of classifiers on spurious background features.
    \item We conduct a comprehensive empirical evaluation on multiple benchmark datasets, spanning diverse training and test settings that vary in the availability of groups (pair of target and spurious attribute) and how the spurious correlation breaks at test time.
    \item We analyze the effect of key design choices and components on the effectiveness of \method{}.
\end{itemize}

\section{Related Work}

\paragraph{Spurious correlations.}
The reliance of neural network predictors on spurious features is a widely known phenomenon \citep{Baker:18, Zech:18, Buolamwini:18, Geirhos:19, Brendel:19, Oakden-Rayner:20} which may lead to detrimental consequences when applied in real world settings. 
\citet{Sagawa:20, Xiao:21, Moayeri:22} emphasize the susceptibility of image classifiers to spurious backgrounds.
One reason for this is the simplicity bias of neural networks \citep{Shah:20, Scimeca:22}, where simple but possibly irrelevant features (also referred to as shortcuts) tend to dominate the prediction.
Comprehensive surveys are given by \citet{Geirhos:20, Yang:23}.

\paragraph{Mitigating reliance on spurious correlations.}
Distributionally robust optimization \citep{Sagawa:20, Levy:20, Sohoni:20, Sohoni:22} aims to learn models that perform well under worst-case conditions, minimizing the maximum risk over all possible groups.
Invariant risk minimization \citep{Arjovsky:19, Ahuja:21, Lin:22} aims to obtain feature representations that are invariant to particular attributes, such that a classifier based on those representations is more robust to shifts in those attributes.
Similarly, correlation alignment \citep{Sun:16a, Sun:16b} encourages invariance by aligning second-order statistics of feature representations across environments, reducing distributional discrepancies caused by spurious attributes.
Domain adversarial neural networks \citep{Ganin:16} learn invariant representations by adding an auxiliary head that predicts the spurious attribute, with its gradient reversed to discourage the shared backbone from encoding it.
Furthermore, there are multiple two-stage approaches, where in the first stage, standard empirical risk minimization (ERM) training is used to obtain a base model.
In the second stage, a new network is trained, upweighting misclassified samples of the base model \citep{Liu:21}, or using contrastive learning to make representations for the same class more similar \citep{Zhang:22}.
\citet{Kirichenko:23} and \citet{Qiu:23} only retrain the last layer, using an additional group-balanced dataset or weighting according to the loss of the base model respectively. 
\citet{Nam:20} also trains two models, one where a spurious bias is amplified which is used by the second one to mitigate the bias.
\citet{Lee:23, Pagliardini:23} learn an ensemble of models with diversity constraints between members, showing that the ensemble becomes more robust to spurious correlations. 
\citet{Yao:22} introduce selective augmentation based on mixup \citep{Zhang:18}, with interpolation between either the same labels with different spurious attributes or vice versa.

\paragraph{Background augmentation.}
Most related to our work are two concurrent methods both termed BackMix: \citet{Bransby:24} randomly sample backgrounds for each foreground under known masks in a semi-supervised echocardiography setting, while \citet{Wang:25} estimate foreground regions via class activation maps and paste background patches onto target images in an open-set recognition setting.
The work most closely related to ours is \citet{Chang:21}. 
They assume access to foreground segmentation masks (or bounding boxes) for data augmentation, constructing factual samples by infilling around a foreground, and counterfactual samples by infilling the background. 
Training combines standard cross-entropy on original data with two additional losses that increase the true-class probability on factual samples and decrease it on counterfactual ones. 
For more details, see Apx.~\ref{sec:detailed_comparison}.

\section{Method}

In this section, we introduce \method{}, a data augmentation approach that reduces classifier reliance on spurious backgrounds. 
We formalize the setting, present a general training algorithm for background-invariant classifiers, and describe our implementation and design choices.

\subsection{Problem Setting}

We consider the classification setting with input images $\Bx \in \cX = \dR^{C\times H \times W}$ and targets $y \in \cY$, where $\cY$ is the set of $K$ possible target classes.
Furthermore, we assume that there is another attribute $a \in \cA$ associated with each input image.
Each possible tuple of $y$ and $a$ forms a group $g$.
In particular, we assume that the foreground $\Bx_{\text{fg}}$ causally determines the target $y$, while the background $\Bx_{\text{bg}}$ causally determines the attribute $a$.
The foreground is defined as $\Bx_{\text{fg}} = \Bx \odot \Bm$, where $\Bm \in \cM = \{0,1\}^{H \times W}$ is a binary mask, and the background is defined as $\Bx_{\text{bg}} = \Bx \odot \left( \bm{1} - \Bm \right)$.
In this setting, spurious correlations arise when background information $\Bx_{\text{bg}}$, defined by attribute $a$, is correlated with target $y$ in the training data despite not being causally related to $y$, often leading to failures when this correlation does not hold at test time.

\subsection{Background Invariance}

We seek to reduce the dependence of a learned classifier $c(\cdot): \cX \rightarrow \cY$ on the background $\Bx_{\text{bg}}$ for a new input $\Bx$.
A direct approach would be to extract and predict using the foreground $\Bx_{\text{fg}}$ only.
However, this requires a robust way of extracting the foreground, both during training and inference, and adds additional inference costs on top of the classifier.
Therefore, we opt for the alternative, training a classifier that is invariant to the background, by means of data augmentation.
The general algorithm is provided in Alg.~\ref{algo} and a visualization of the main steps in Fig.~\ref{fig:method}.
It assumes access to a dataset $\cD = \{(\Bx_i, y_i, a_i)\}_{i=1}^N$, a detector $d(\cdot): \cX \rightarrow \cM$ and a generator $g(\cdot): \cX \rightarrow \cX$.
The detector predicts the binary mask $\Bm$ that defines the foreground and background, given an input image.
The generator performs infilling, producing a full background image given the background image after masking the foreground.
Finally, for creating a new augmented sample for training, a foreground is pasted onto a new infilled background.
The background is randomly selected from the set of backgrounds that exhibit the spurious attribute, where the spurious attribute is randomly selected each time as well, breaking the correlation between target and spurious attribute in the input images.

\subsection{Implementing \method{}} \label{sec:autobackswap}

The general algorithm in Alg.~\ref{algo} took the detector $d(\cdot)$ and the generator $g(\cdot)$ as given.
In the following, we provide details on how we implement them.
Furthermore, the algorithm assumed access to the attribute $a$ for each sample.
Although optimal if available, we do not consider them as given in our experiments and discuss viable approximations, making \method{} applicable in a more broad range of settings.

\begin{wrapfigure}{R}{0.55\textwidth}
\resizebox{0.55\textwidth}{!}{
\begin{minipage}{0.6\textwidth}
\vspace{-0.4cm}
\begin{algorithm}[H]
\caption{Generalized \method{}}
\label{algo}
\begin{algorithmic}
    \STATE {\bfseries Input:} Dataset $\cD$, detector $d(\cdot)$, generator $g(\cdot)$ \\
    \comm{Generate disentangled fore- and background datasets}
    \STATE $\cF \gets \emptyset$, $\cB \gets \emptyset$
    \FORALL{$(\Bx, y, a) \in \cD$}
        \STATE $\Bm \gets d(\Bx)$ \com{detect foreground}
        \STATE $\Bx_{\text{fg}} \gets \Bx \odot \Bm$ \com{apply mask for foreground}
        \STATE $\Bx_{\text{bg}} \gets \Bx \odot \left( \bm{1} - \Bm \right)$ \com{apply mask for background}
        \STATE $\Bx_{\text{bg}} \gets g(\Bx_{\text{bg}})$ \com{inpaint background}
        \STATE $\cF \gets \cF \cup \{ (\Bx_{\text{fg}}, y) \}$
        \STATE $\cB \gets \cB \cup \{ (\Bx_{\text{bg}}, a) \}$
    \ENDFOR \\
    \comm{Train classifier on augmented samples}
    \STATE Initialize classifier $c(\cdot)$
    \WHILE{not converged}
        \STATE $(\Bx_{\text{fg}}, y) \sim \text{Unif}(\cF)$ \com{sample foreground and target}
        \STATE $a \sim \text{Unif}(\cA)$ \com{sample spurious attribute}
        \STATE $\cB_a \gets \{\, \Bx_{\text{bg}} \mid (\Bx_{\text{bg}}, a)\in \cB\,\}$
        \STATE $\Bx_{\text{bg}} \sim \text{Unif}(\cB_{a})$ \com{sample background acc. to attribute}
        \STATE $\tilde{\Bx} \gets \text{Paste}(\Bx_{\text{fg}}, \Bx_{\text{bg}})$ \com{paste fore- on background}
        \STATE Update $c(\cdot)$ by ERM step on (a minibatch of) $(\tilde{\Bx}, y)$
    \ENDWHILE
    \STATE {\bfseries Return:} classifier $c(\cdot)$.
\end{algorithmic}
\end{algorithm}
\vspace{-1.0cm}
\end{minipage}}
\end{wrapfigure}

\paragraph{Detector.}
We consider a secondary model, typically of a similar structure as the classifier, on the binary classification task foreground vs. background on a patchwise level.
For this, we assume that an auxiliary dataset $\cD_{\text{aux}} = \{(\Bx_i, \Bz_i)\}_{i=1}^M$ with patch-wise labels $\Bz \in \{0, 1\}^{{R}, {R}}$ is available. 
The resolution $R$ is an important parameter. 
Too fine-grained and the task becomes hard to learn, requiring more instances $M$ and thus more effort to create the auxiliary dataset.
Too coarse and the separation of foreground and background becomes too imprecise to be useful.
This tradeoff is investigated in more detail in Sec.~\ref{sec:ablation2}.
We find that very few examples are often sufficient to train a detector with high accuracy on the classifier training data $\cD$.
More details on the implementation of the detector in our experiments are provided in Apx.~\ref{sec:apx:our_method}.

Zero-shot segmentation models such as SAM3 \citep{Carion:25} can be a powerful alternative to training a dedicated model on additional data.
However, zero-shot performance could be rather suboptimal on, e.g. industrial or scientific tasks that are very dissimilar to the models training distribution.
Furthermore, general-purpose models such as SAM3 are generally much larger than specialised models (about 30 times in our experimental setup), incurring higher inference costs for the entire pipeline.

\paragraph{Generator.}
For background infilling, we use the AOT-GAN model \citep{Zeng:23} as it provides a favorable tradeoff between runtime and infilling quality. In our experiments in Sec.~\ref{sec:ablation1}, we find that infilling is often not the dominant factor for performance, and even simple pixel scrambling or no infilling at all can yield strong results. We therefore choose AOT-GAN as an efficient default option with competitive inpainting quality. While diffusion-based inpainting models are also possible and often provide slightly better visual quality than GAN-based approaches, they are substantially more computationally demanding, since they require dozens of iterative denoising steps for each image.

\paragraph{Recomposing $\Bx_{\text{fg}}$ and $\Bx_{\text{bg}}$.}
A foreground image $\Bx_{\text{fg}}$ is pasted onto a sampled inpainted background image $\Bx_{\text{bg}}$. 
We find that additional augmentation of the foreground further boosts the performance of \method{}.
In particular, we consider resizing, translation and rotation as foreground augmentations before pasting onto the background.

\paragraph{Access to spurious attributes.}
It may often be an inconvenient requirement to have access to information about the spurious attribute in the training data.
Throughout our experiments, we drop this requirement and instead sample according to the target class $y \sim \text{Unif}(\cY)$ instead of the spurious attribute and then sampling the background from the subset $\cB_y = \{ \Bx_{\text{bg}} \mid (\Bx_{\text{bg}}, y) \in \cB \}$.
This is a good approximation to Alg.~\ref{algo} assuming that the target class is strongly correlated with the spurious attribute in the training data, which is why we worry about it in the first place.

\section{Experiments} \label{sec:exp}

In this section, we investigate the performance of \method{} on different benchmarks.
First, the widely considered Waterbirds dataset \citep{Sagawa:20}, which features both binary targets $\cY$ and spurious attributes $\cA$, artificially constructed by combining a target-indicative foreground with a potentially spurious background.
We consider two settings: one in which minority groups are available during training and one in which they are not, see Fig.~\ref{fig:waterbirds}.
Second, the Spawrious dataset \citep{Lynch:23} features four targets and six backgrounds, created using a text-to-image diffusion model, and supports one-to-one (o2o) and many-to-many (m2m) experimental settings, see Fig.~\ref{fig:spawrious}.
Third, we propose the Spurious Vehicles dataset, featuring four vehicle classes in four contextual environments, evaluated under the m2m setting, see Fig.~\ref{fig:spurious_vehicles_configurations}.

\subsection{Experimental Setup} \label{sec:setup}

\paragraph{Baseline methods.}
We consider the following methods as baselines:
\textbf{(1)} Standard ERM
\textbf{(2)} ERM with heavy augmentation (mixup \citep{Zhang:18}, CutMix \citep{Yun:19}, and ColorJitter)
\textbf{(3)}~DFR \citep{Kirichenko:23}
\textbf{(4)} AFR \citep{Qiu:23}
\textbf{(5)} GroupDRO \citep{Sagawa:20} 
\textbf{(6)} CORAL \citep{Sun:16a,Sun:16b}
\textbf{(7)} the method by \citet{Chang:21}.

Furthermore, we consider the following three foundation models, which can be used as classifiers in a zero-shot manner:
\textbf{(8)} SAM 3 \citep{Carion:25}
\textbf{(9)} CLIP \citep{Radford:21}
\textbf{(10)}~QWEN3-VL-Thinking \citep{Bai:25}.
We include these baselines, to assess whether large-scale foundation models are also biased towards background information for performing zero-shot predictions.
Additional details on all baseline methods are provided in Apx.~\ref{sec:apx:baselines}.

\paragraph{Models.}
Following previous works \citep{Sagawa:20, Kirichenko:23, Qiu:23, Arefin:24} the results provided in the main paper are based on pretrained ResNet50 models \citep{He:16}.
We investigate the reliability of the obtained results under a change to pretrained ViT models \citep{Dosovitskiy:21} in Apx.~\ref{sec:apx:waterbirds_vit}.

\paragraph{Metrics.}
We use the standard accuracy \textit{Acc} on the full dataset and the worst group accuracy $\text{\textit{WGA}} = \min_{g} \text{\textit{Acc}}_{g}$ as evaluation metrics.
We depart from the common practice of \citet{Sagawa:20}, which reports group-weighted accuracy with weights proportional to group prevalence in the training data.
Although such weighting can be appropriate when the same groups appear in both training and test sets, it is not suited to our experiments where groups differ.

\paragraph{Information about spurious attribute.}
We use the standard accuracy on the validation dataset for early stopping for all considered methods to avoid using information about the spurious attribute during training.
Early stopping is used by all methods (except AFR, see Apx.~\ref{sec:apx:baselines} for details) for model selection.
Noteworthy, the baseline methods Group DRO and CORAL need group information for loss calculations; DFR also implicitly as a group-balanced reweighting dataset is used in the second stage.
The method by \citet{Chang:21} assumes access to ground-truth foreground segmentations for the full training dataset.
For the Waterbirds dataset, they are provided; for the different Spawrious tasks we use SAM3 to obtain them, see Apx.~\ref{sec:apx:baselines} for details.
QWEN3-VL-Thinking is prompted to ignore the background.
\method{} learns the detector model based on a few hundred auxiliary samples.

\subsection{Waterbirds} \label{sec:waterbirds}

\paragraph{Setup.}
The Waterbirds dataset \citep{Sagawa:20} consists of images of waterbirds and, pasted onto background images showing either water or land.
Thus the class labels are $\cY$~=~\{waterbird, landbird\} and the spurious attribute is $\cA$~=~\{water background, land background\}, leading to four different groups, see Fig.~\ref{fig:waterbirds}.
Groups $g_1$ and $g_4$ (landbirds on land, waterbirds on water) dominate the training set and are referred to as majority groups, while $g_2$ and $g_3$ (landbirds on water, waterbirds on land) are the minority groups.
We consider two settings in our experiments:
\textit{(i)} training on the full dataset, and \textit{(ii)} training only on majority groups.
In the training set, groups have different prevalence, containing 3498, 184, 56 and 1057 samples respectively.

\begin{wrapfigure}{R}{0.42\textwidth}
\vspace{-0.42cm}
\setlength{\tabcolsep}{0.5pt}
\renewcommand{\arraystretch}{0.95}
\centering
\captionof{table}{Results for Waterbirds experiments, training either with or without minority groups. Best result \textbf{bold}, second best \underline{underlined}. Statistics computed over five independent runs.\vspace{-0.2cm}}
\label{tab:waterbirds_main}
\resizebox{0.42\textwidth}{!}{
\begin{tabular}{ccccccc}
\toprule
\multirow{2}{*}{\textbf{Method}} && \multicolumn{2}{c}{\textbf{w minority}} && \multicolumn{2}{c}{\textbf{w/o minority}} \\ \cline{3-4}\cline{6-7}
\textbf{}                         &&  \textit{WGA}           & \textit{Acc}           &&  \textit{WGA}            & \textit{Acc}       \\ \midrule
ERM                               &&           $74.9_{(3.3)}$           &           $93.6_{(0.4)}$      &&          $33.4_{(4.4)}$           &      $68.4_{(1.9)}$                \\
+ Heavy Aug.                   &&       $74.3_{(2.3)}$              &           $87.4_{(1.0)}$         &&            $15.9_{(2.7)}$          &          $61.5_{(1.2)}$          \\
DFR                               &&           $\underline{90.7}_{(0.8)}$          &         $94.3_{(0.1)}$           &&         \textcolor{black}{$15.1_{(1.9)}$}             &           \textcolor{black}{$60.1_{(0.8)}$}        \\
AFR                               &&             $78.6_{(3.7)}$        &              $88.5_{(0.9)}$      &&             \textcolor{black}{$22.2_{(3.3)}$}         &           \textcolor{black}{$64.2_{(0.8)}$}         \\
Group DRO                         &&            $82.2_{(1.6)}$          &         $94.8_{(0.2)}$           &&            $27.6_{(3.8)}$          &           $65.3_{(0.6)}$         \\
CORAL                             &&             $78.4_{(4.2)}$        &          $93.6_{(0.3)}$          &&           $22.0_{(5.5)}$           &            $65.0_{(2.3)}$        \\
Chang et al. &&          $85.7_{(2.3)}$           &           $\boldsymbol{95.7}_{(0.3)}$         &&           $\underline{79.1}_{(3.2)}$           &           $\underline{91.4}_{(1.8)}$         \\ 
\rowcolor{gray!15} \textbf{Ours}        &&           $\boldsymbol{93.0}_{(0.9)}$          &         $\underline{95.0}_{(0.5)}$           &&            $\boldsymbol{92.2}_{(0.4)}$           &              $\boldsymbol{95.1}_{(0.3)}$  \\ \hdashline \\ [-2.1ex]
SAM3                              &&          $38.0$          &          $64.1$          &&            $38.0$          &           $64.1$         \\
CLIP                              &&           $31.5$          &           $60.0$         &&           $31.5$           &           $60.0$         \\
QWEN3-VL &&          $70.6$           &           $86.5$         &&          $70.6$            &          $86.5$          \\ \hline
\end{tabular}}
\vspace{-0.5cm}
\renewcommand{\arraystretch}{1}
\end{wrapfigure}

\paragraph{Results.}
The results for both settings are provided in Tab.~\ref{tab:waterbirds_main}.
For the first setting \textit{(i)} with minority groups, we find that many of the considered baseline methods significantly outperform the naive ERM baseline.
\method{} has the highest WGA in this setting by more than 2\%, and is second best in Acc, trailing by 0.7\%.
DFR has the second best WGA, while \citet{Chang:21} has the highest Acc.

For the second, more challenging setting \textit{(ii)} without minority groups, the performance of all baselines except \citet{Chang:21} drops drastically.
This is expected, as their assumptions are broken when no minority samples that break the correlation between $\cY$ and $\cA$ are available.
\method{} strongly outperforms all baselines in this setting, improving \textbf{13.1\%} in WGA and \textbf{3.7\%} in Acc, compared to the next-best baseline.

As expected, the strongest baseline across both settings is \citet{Chang:21}.
Their method uses ground-truth foreground / background segmentations on a pixel level to generate additional factual and counterfactual data.
\method{} is learning a foreground / background detector based on 287 auxiliary samples with much coarser patch-level information to generate augmented data.

Contrary to our expectations, ERM with heavy augmentation underperforms ERM with the standard augmentations we used across all methods (horizontal flipping and random resized crop), across both settings for both WGA and Acc.
The validation accuracy of ERM with and without heavy augmentation is comparable, ruling out underfitting as an explanation.
We hypothesize that the considered heavy augmentations reinforce the classifiers' bias towards background information.
Studying this effect in more detail would be an interesting direction for future work.

Among the zero-shot foundation models, SAM3 and CLIP underperform even the naive ERM baseline. 
QWEN3-VL-Thinking is more competitive, benefiting from its ability to incorporate knowledge that the background is spuriously correlated into its prediction through prompting.
Without such prompting, its performance drops significantly.

\subsection{Spawrious} \label{sec:spawrious}

\paragraph{Setup.}
The Spawrious dataset \citep{Lynch:23} consists of images of four dog breeds placed on six different background settings. 
The target classes are $\cY$~=~\{bulldog, dachshund, labrador, corgi\}, and the spurious background attributes are $\cA$~=~\{desert, jungle, dirt, snow, beach, mountain\}, yielding 24 possible target–background group combinations. 
The benchmark defines two experimental settings, \textit{one-to-one} and \textit{many-to-many}, each further divided into three difficulty levels, \textit{easy}, \textit{medium} and \textit{hard}, which differ in the specific groups included in the training and test sets. 
Fig.~\ref{fig:spawrious} in the appendix provides the exact group configurations for each setting and difficulty level.

In the \textit{one-to-one} setting, each target class $y$ is associated with a dominant spurious background attribute $a$ and a single shared background across all classes. 
The training data follow a 9:1 imbalance, with the minority samples drawn from the shared background to partially break the correlation between $y$ and $a$. 
At test time, the spurious attribute assigned to each target is shifted, yielding target–background combinations unseen during training.

In the \textit{many-to-many} setting, each target class is associated with two background attributes in equal proportion, with two target classes sharing the same backgrounds. 
For testing, the target–background assignments are swapped, so that the same targets still share backgrounds, but with reversed associations compared to training.

\setlength{\tabcolsep}{0.2pt}
\renewcommand{\arraystretch}{1.1}
\begin{table}
\centering
\caption{Results for Spawrious one-to-one and many-to-many tasks. We report \textit{WGA} and \textit{Acc} in the respective test groups. Best result \textbf{bold}, second best \underline{underlined}; zero-shot baselines are excluded from comparison and marked with $^*$ if best. Statistics computed over five runs.\vspace{0.1cm}}
\label{tab:spawrious}
\small
\resizebox{\textwidth}{!}{
\begin{tabular}{ccccccccccccccccccc}
\toprule
\multirow{2}{*}{\textbf{Method}}  && \multicolumn{2}{c}{\textbf{o2o-easy}}  && \multicolumn{2}{c}{\textbf{o2o-medium}}  &&  \multicolumn{2}{c}{\textbf{o2o-hard}} && \multicolumn{2}{c}{\textbf{m2m-easy}}  && \multicolumn{2}{c}{\textbf{m2m-medium}}  &&  \multicolumn{2}{c}{\textbf{m2m-hard}}  \\ \cline{3-4} \cline{6-7} \cline{9-10} \cline{12-13} \cline{15-16} \cline{18-19}
                                  && \textit{WGA} & \textit{Acc} && \textit{WGA} & \textit{Acc} && \multicolumn{1}{c}{\textit{WGA}} & \multicolumn{1}{c}{\textit{Acc}} && \multicolumn{1}{c}{\textit{WGA}} & \multicolumn{1}{c}{\textit{Acc}} && \multicolumn{1}{c}{\textit{WGA}} & \multicolumn{1}{c}{\textit{Acc}} && \multicolumn{1}{c}{\textit{WGA}} & \multicolumn{1}{c}{\textit{Acc}} \\ \midrule
ERM                               &&           $89.1_{(1.1)}$          &           $93.3_{(0.5)}$         &&           $34.5_{(2.5)}$          &          $80.8_{(0.9)}$          &&                     $71.7_{(2.2)}$                    &           $87.9_{(0.5)}$                             &&                     $81.1_{(1.4)}$                    &                   $93.0_{(0.4)}$                     &&              $50.7_{(5.0)}$                           &                     $80.4_{(1.7)}$                   &&                     $51.3_{(3.9)}$                    &                $74.9_{(1.1)}$                        \\
+ Heavy Aug.                   &&            $31.6_{(6.7)}$         &             $45.8_{(4.3)}$       &&           $11.3_{(2.4)}$          &          $67.5_{(1.3)}$          &&                   $25.9_{(3.2)}$                      &                   $52.3_{(4.8)}$                     &&                    $50.9_{(9.5)}$                     &                    $79.2_{(3.7)}$                    &&                  $28.5_{(5.9)}$                       &                      $58.4_{(4.7)}$                  &&                     $19.6_{(10.8)}$                    &                 $45.2_{(14.7)}$                       \\
DFR                               &&           $89.6_{(1.4)}$          &             $94.1_{(0.4)}$       &&           $45.2_{(2.6)}$          &           $83.8_{(0.7)}$         &&                    $64.5_{(10.4)}$                     &                  $86.5_{(2.9)}$                      &&                   $75.8_{(1.0)}$                      &                     $91.0_{(0.4)}$                   &&              $49.8_{(1.3)}$                           &                      $80.7_{(0.6)}$                  &&                      $39.1_{(4.1)}$                   &              $67.7_{(4.6)}$                          \\
AFR                               &&           $83.8_{(1.2)}$          &             $89.8_{(0.5)}$       &&           $31.7_{(1.7)}$          &           $78.1_{(0.4)}$         &&                   $54.7_{(3.0)}$                      &                    $79.6_{(0.8)}$                    &&                     $71.4_{(1.3)}$                    &                     $90.3_{(0.5)}$                   &&                  $40.4_{(3.0)}$                       &                     $73.9_{(0.8)}$                  &&                      $41.3_{(1.2)}$                   &                  $68.5_{(1.2)}$                      \\
Group DRO                         &&             $\underline{91.9}_{(0.3)}$         &          $95.1_{(0.2)}$          &&            $54.9_{(1.9)}$         &         $86.9_{(0.4)}$           &&                   $\underline{80.1}_{(1.5)}$                      &                 $90.5_{(0.2)}$                       &&                     $77.6_{(2.4)}$                    &                    $92.2_{(0.7)}$                    &&                 $47.3_{(4.0)}$                        &                      $79.0_{(1.2)}$                  &&                       $47.1_{(5.7)}$                  &                           $71.5_{(2.9)}$             \\
CORAL                             &&           $91.3_{(1.7)}$          &           $94.7_{(0.5)}$         &&          $43.0_{(5.4)}$           &           $83.1_{(1.2)}$         &&                   $71.0_{(3.9)}$                      &                $88.6_{(1.0)}$                        &&                       $81.4_{(1.2)}$                  &                  $93.2_{(0.2)}$                      &&                 $54.0_{(2.3)}$                         &                      $81.7_{(0.8)}$                  &&                     $49.8_{(5.0)}$                    &                 $74.0_{(2.0)}$                       \\ 
Chang et al. &&             $\boldsymbol{93.3}_{(0.6)}$        &          $\boldsymbol{96.9}_{(0.1)}$          &&          $\underline{70.4}_{(3.1)}$           &           $\underline{91.2}_{(0.7)}$         &&                     $74.9_{(2.3)}$                    &                   $\underline{90.9}_{(0.4)}$                     &&                      $\underline{89.5}_{(1.5)}$                   &             $\underline{96.4}_{(0.3)}$                           &&                     $\underline{77.8}_{(1.9)}$                    &                    $\underline{90.8}_{(0.3)}$                    &&                  $\boldsymbol{66.7}_{(1.7)}$                       &                      $\boldsymbol{85.7}_{(0.2)}$                  \\ 
\rowcolor{gray!15}\textbf{Ours}        &&           $88.9_{(1.7)}$          &          $\underline{95.8}_{(0.3)}$          &&           $\boldsymbol{72.5}_{(4.0)}$          &           $\boldsymbol{91.5}_{(1.0)}$         &&                   $\boldsymbol{81.4}_{(2.2)}$                      &                 $\boldsymbol{92.7}_{(0.6)}$                       &&                      $\boldsymbol{92.6}_{(0.5)}$                   &                         $\boldsymbol{97.0}_{(0.1)}$               &&                     $\boldsymbol{78.0}_{(1.5)}$                    &                        $\boldsymbol{91.2}_{(0.2)}$                &&                          $\underline{64.3}_{(6.6)}$               &                    $\boldsymbol{85.7}_{(0.8)}$                    \\ \hdashline \\[-2.1ex]
SAM3                              &&          $59.9$           &          $79.6$          &&           $51.4$          &            $77.2$        &&                     $59.9$                    &                  $79.3$                      &&                     $59.7$                    &                $80.6$                        &&                     $51.2$                    &                     $77.6$                   &&                    $64.0$                     &                   $81.6$                     \\
CLIP                              &&        $79.6$             &            $93.4$        &&         $\phantom{^*}90.3^*$            &           $\phantom{^*}95.5^*$         &&            $79.6$                             &                    $91.9$                    &&                  $92.4$                       &                   $\phantom{^*}97.0^*$                     &&                    $\phantom{^*}87.2^*$                     &                   $\phantom{^*}94.0^*$                     &&                         $\phantom{^*}79.9^*$                &                    $\phantom{^*}92.8^*$                    \\
QWEN3-VL   &&           $72.8$          &          $87.5$          &&          $\phantom{^*}85.7^*$            &           $90.3$           &&                 $73.4$                        &                     $86.8$                   &&                    $85.7$                     &                       $93.0$                 &&                     $\phantom{^*}81.8^*$                    &                       $89.5$                 &&                  $\phantom{^*}73.8^*$                       &                     $\phantom{^*}88.9^*$                   \\ \hline
\end{tabular}
}
\vspace{-0.3cm}
\setlength{\tabcolsep}{1.3pt}

\end{table}
\renewcommand{\arraystretch}{1}

\paragraph{Results.}
The results for both settings and all difficulty levels are reported in Tab.~\ref{tab:spawrious}. 
\method{} achieves the best performance in nine out of twelve experimental conditions (e.g., WGA in o2o-hard) and ranks second in two additional cases. 
Among the competing approaches, only \citet{Chang:21} attains comparable performance, ranking first in four and second in seven conditions. 
Of the remaining baselines, GroupDRO is the only method to place among the top performers, achieving second-best results in two conditions.

In the one-to-one setting, all methods perform similarly due to the shared background, which partially breaks the correlation between the target $y$ and the spurious attribute $a$.
This partial decorrelation satisfies a key implicit assumption of DFR, AFR, GroupDRO, and CORAL, enabling these methods to perform well. 
In contrast, in the many-to-many setting, the correlation remains largely intact, resulting in substantially degraded performance for these approaches. 
Only \citet{Chang:21} and \method{} are explicitly designed to handle this regime by leveraging additional supervision to disentangle $y$ and $a$.
Specifically, \citet{Chang:21} relies on ground-truth foreground segmentation masks, while \method{} employs 400 auxiliary patch-level labels to train the detector.
Finally, heavy data augmentation consistently degrades performance, even more than for Waterbirds.

Finally, several foundation models perform competitively with task-specific methods.
CLIP achieves the strongest overall results, followed by QWEN3-VL-Thinking.
The experimental conditions where they outperform the best task-specific model are marked with a star in Tab.~\ref{tab:spawrious}.
We attribute the good performance of foundation models to the prevalence of dog images in large-scale pretraining data, making the Spawrious tasks rather in-distribution, and to the absence of strong background–target correlations in these tasks, in contrast to Waterbirds.

\subsection{Spurious Vehicles} \label{sec:spurious_vehicles}

\paragraph{Setup.}
The \textit{Spurious Vehicles} dataset is a synthetic benchmark proposed in this work to study spurious correlations in a vehicle recognition setting. It consists of images of four vehicle classes placed in four contextual environments. The target classes are $\cY=\{\text{sedan}, \text{minivan}, \text{SUV}, \text{pickup truck}\}$ and the spurious context attributes are $\cA=\{\text{urban}, \text{highway}, \text{rural}, \text{off-road}\}$, yielding 16 possible target-context group combinations. Example images and the exact train/test group assignments are provided in the appendix.
We use the m2m setting, analogous to Spawrious.

\begin{wrapfigure}{R}{0.38\textwidth}
\vspace{-0.45cm}
\setlength{\tabcolsep}{1.8pt}
\renewcommand{\arraystretch}{1}
\centering
\captionof{table}{Results for the Spurious Vehicles experiments. Best result \textbf{bold}, second best \underline{underlined}. Statistics computed over three independent runs.\vspace{-0.2cm}}
\label{tab:vehicles_main}
\resizebox{0.38\textwidth}{!}{
\begin{tabular}{ccc}
\toprule
\textbf{Method} & \textit{WGA} & \textit{Acc} \\ 
\midrule
ERM                  & $9.5_{(11.1)}$               & $37.9_{(5.6)}$ \\
+~Heavy Aug.             & $2.1_{(2.9)}$                & $11.8_{(4.1)}$ \\
DFR                  & $9.2_{(4.9)}$                & $54.9_{(5.6)}$ \\
AFR                  & $0.1_{(0.2)}$                & $17.9_{(5.0)}$ \\
Group DRO            & $4.8_{(4.1)}$                & $36.9_{(9.2)}$ \\
CORAL                & $4.7_{(6.8)}$                & $27.8_{(3.9)}$ \\
Chang et al.         & $\underline{87.5}_{(2.7)}$  & $\underline{95.3}_{(0.6)}$ \\
\rowcolor{gray!15} \textbf{Ours (GT Seg.)} & $\boldsymbol{90.1}_{(2.5)}$ & $\boldsymbol{96.3}_{(0.6)}$ \\
\rowcolor{gray!15} \textbf{Ours (Learned Det.)}         & $87.2_{(1.9)}$          & $\underline{95.3}_{(0.7)}$ \\
\bottomrule
\end{tabular}}
\renewcommand{\arraystretch}{1}
\vspace{-0.4cm}
\end{wrapfigure}

\paragraph{Results.}
The results for the Spurious Vehicles benchmark are reported in Tab.~\ref{tab:vehicles_main}. We observe that most standard baselines fail almost completely in this setting, indicating that it poses a particularly challenging spurious correlation scenario where conventional robustness methods struggle to generalize.
Among the baselines, only \citet{Chang:21}, which assumes access to ground-truth segmentation, achieves strong performance. Our method with the learned lightweight detector reaches a highly competitive worst-group accuracy while matching the best baseline in overall accuracy. This demonstrates that effective mitigation is possible without requiring ground-truth segmentations during training.

When replacing the learned detector with ground-truth segmentations (GT Seg.), performance further improves and achieves the best overall results. This suggests that the remaining gap is primarily due to foreground/background separation quality rather than limitations of the proposed augmentation pipeline itself. Overall, the results show that \method{} remains highly effective even with a lightweight learned detector, while improved segmentations can yield additional gains.

\section{Ablations on Key Design Choices} \label{sec:exp:design_choices}

In this section, we provide an in-depth analysis of design choices on key aspects of the three main steps of \method{}, foreground detection, background infilling and sample recomposition.

\subsection{Quality of Auxiliary Dataset} \label{sec:ablation2}

\begin{figure}[t!]
    \centering
    \hfill
    \includegraphics[width=0.48\textwidth]{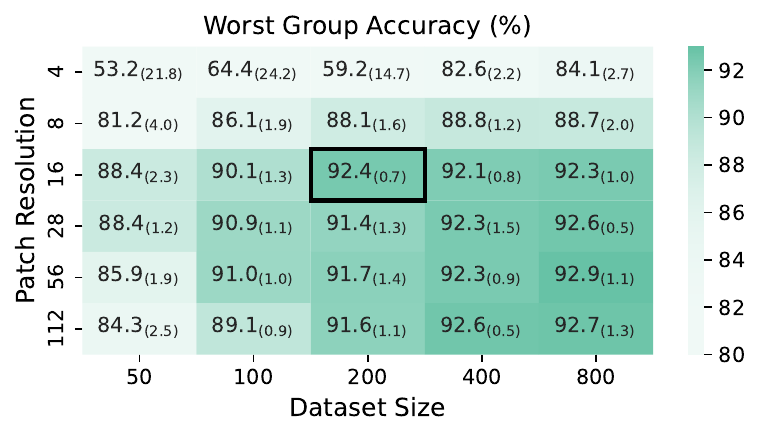}
    \hfill
    \includegraphics[width=0.485\textwidth]{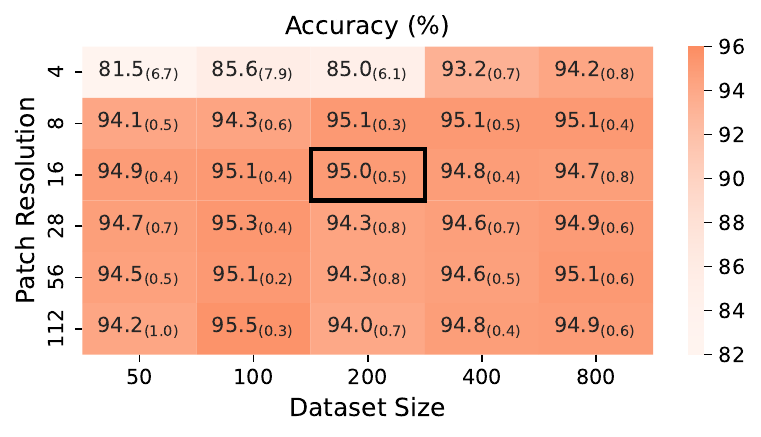}
    \hfill
    \vspace{-0.2cm}
    \caption{Dependency of \method{}'s performance on dataset size and patch resolution for Waterbirds (w/o minority). Both higher patch resolution and larger dataset size imply higher costs due to increased labeling effort. The black border denotes the setting closest to our main experiments (patch size 16, dataset size 287), striking a favorable tradeoff between performance and costs.}
    \label{fig:ablation_detector_quality}
    \vspace{-0.3cm}
\end{figure}

\method{} uses an auxiliary dataset with patch-wise foreground mask labels to train the detector.
Each mask $\Bz \in \{0, 1\}^{{R}, {R}}$ is provided at a fixed patch resolution $R$, where higher resolution means more patches (see Fig.~\ref{fig:ablation_detector_quality_examples} for example images).
Choosing an appropriate resolution involves a tradeoff: coarse masks provide little utility for disentangling foreground and background with the detector, while fine-grained masks are harder to predict and require more labeled data, increasing annotation cost.
Therefore, we study how patch resolution and dataset size jointly affect performance.

\begin{wrapfigure}{r}{0.55\linewidth}
    \vspace{-0.3cm}
    \includegraphics[width=1.\linewidth]{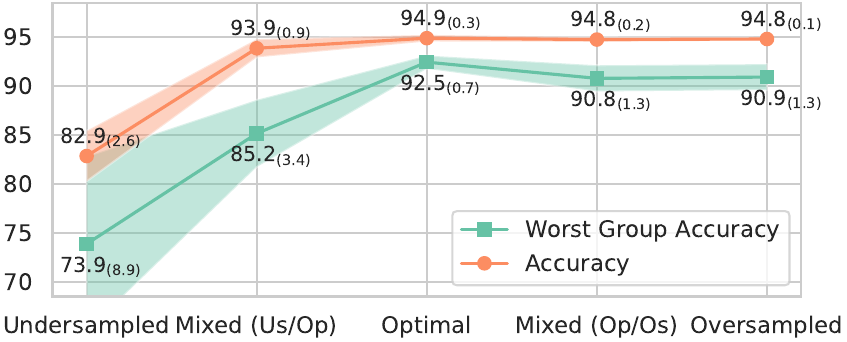}
    \caption{Effect of biased patch-wise foreground-mask labels on \method{}'s performance. We compare \textit{undersampling}, \textit{optimal}, \textit{oversampling}, and mixed label settings. Performance is robust to oversampling but degrades sharply under undersampling.}
    \label{fig:ablation_labels}
    \vspace{1cm}
    \includegraphics[width=1.0\linewidth]{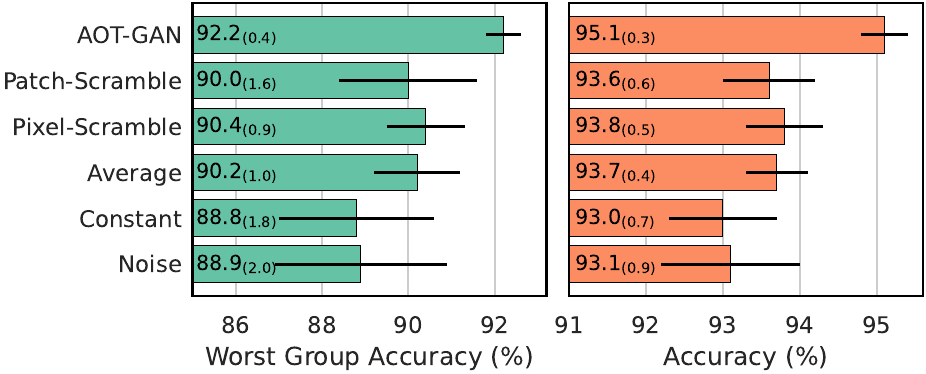}
    \caption{Ablation on different infilling variants. Using AOT-GAN as in our main experiments results in the best performance, yet also more simple infilling methods lead to competitive results with only modest drops in WGA and Acc.}
    \label{fig:ablation_inpainting}
\vspace{-1cm}
\end{wrapfigure}

Results in Fig.~\ref{fig:ablation_detector_quality}, obtained on Waterbirds w/o minority (Sec.~\ref{sec:waterbirds}), show that resolutions of 8 and 4 lead to substantial WGA drops across all dataset sizes.
Increasing the patch resolution above 16 yield modest gains in WGA, if the dataset size is large enough.
Accuracy remains largely stable, except for very small resolutions and dataset sizes.
Overall, the configuration considered in the main experiments (highlighted by black border) strikes a favorable tradeoff between performance and labeling costs.

\subsection{Biased Auxiliary Mask Labels} \label{sec:ablation3}

Next, we investigate the effect of biased and noisy auxiliary mask labels on the performance of \method{}.
Using the Waterbirds ground-truth segmentations, we construct auxiliary datasets of size 287 at patch resolution 16.
We evaluate three labeling strategies: \textit{undersampling}, where only patches fully containing the bird are labeled foreground; \textit{oversampling}, where a single foreground pixel suffices; and \textit{optimal} sampling, which labels patches as foreground if at least 25\% of pixels belong to the bird, chosen to closely match our manual annotations (Sec.~\ref{sec:waterbirds}).
We also test stochastic mixtures of the masks obtained through undersampling and optimal, and through optimal and oversampling, selecting from either of them with equal probability, see Fig.~\ref{fig:ablation_label_quality_examples} in the appendix for examples.
Results in Fig.~\ref{fig:ablation_labels} show that \method{} is robust to oversampling, with only a slight drop in WGA compared to optimal sampling.
In contrast, undersampling causes substantial degradation in both WGA and Acc, with a more pronounced impact on WGA.

\subsection{Background Infilling Strategy} \label{sec:ablation1}

The second step in \method{} is background infilling, implemented using the AOT-GAN model \citep{Zeng:23} in the main experiments.
However, suitable pre-trained models may not be available for all background types.
We therefore evaluate the performance of \method{} under simpler infilling strategies based on image statistics and noise.
Specifically, we consider infilling by patch scrambling, pixel scrambling, the per-image average pixel value, a constant value across images, and Gaussian noise (see Fig.~\ref{fig:waterbirds_infill_ablation_examples} for examples).

The results in Fig.~\ref{fig:ablation_inpainting} show that AOT-GAN achieves the best performance, but only by a modest margin of approximately 2\% in WGA and 1\% in Acc.
Infilling methods based on image statistics, i.e. patch-scrambling, pixel-scrambling and the average pixel value, perform comparably to each other, close to that of AOT-GAN.
In contrast, using a constant value or Gaussian noise yields the weakest results.
Overall, these findings indicate that background infilling has a limited impact, and even simple methods such as the average pixel value can yield competitive performance.

\subsection{Sampling Strategy for Recomposition}

\begin{wrapfigure}{R}{0.46\textwidth}
\vspace{-0.42cm}
\setlength{\tabcolsep}{0.6pt}
\renewcommand{\arraystretch}{1.}
\centering
\captionof{table}{Ablation for sampling strategy on Waterbirds dataset. The second strategy (grey) is used throughout all other experiments.\vspace{-0.2cm}}
\label{tab:sampling_strategy}
\resizebox{0.46\textwidth}{!}{
\begin{tabular}{ccccccc}
\toprule
\textbf{Sampling} && \multicolumn{2}{c}{\textbf{w minority}} && \multicolumn{2}{c}{\textbf{w/o minority}} \\ \cline{3-4}\cline{6-7}
\textbf{Strategy}                         &&  \textit{WGA}           & \textit{Acc}           &&  \textit{WGA}            & \textit{Acc}       \\ \midrule
\begin{tabular}{c} $a \sim \text{Unif}(\cA)$\\ $\Bx_\text{bg} \sim\text{Unif}(\cB_a)$ \end{tabular}        &&           $93.8_{(0.4)}$         &          $95.5_{(0.2)}$         &&            $92.2_{(0.9)}$           &         $95.1_{(0.4)}$       \\ 
\rowcolor{gray!15}  \begin{tabular}{c} \rowcolor{gray!15} $y \sim \text{Unif}(\cY)$\\ \rowcolor{gray!15} $\Bx_\text{bg} \sim\text{Unif}(\cB_y)$ \end{tabular}      &&           $93.0_{(0.9)}$          &         $95.0_{(0.5)}$           &&            $92.2_{(0.4)}$           &              $95.1_{(0.3)}$  \\ 
$\Bx_\text{bg} \sim \text{Unif}(\cB)$        &&            $93.1_{(0.6)}$         &          $95.6_{(0.5)}$         &&           $91.7_{(0.8)}$            &              $95.1_{(0.3)}$  \\ \bottomrule
\end{tabular}}
\renewcommand{\arraystretch}{1}
\vspace{-0.3cm}
\end{wrapfigure}

Finally, we study the role of the sampling strategy in breaking the correlation between the target $y$ and the spurious attribute $a$.
Given a foreground $\Bx_{\text{fg}}$ exhibiting some target $y$, the spurious attribute should be sampled uniformly, i.e. $a \sim \text{Unif}(\cA)$, based on which the set of backgrounds exhibiting this attribute $\cB_a$ is selected and a background $\Bx_{\text{bg}} \sim \text{Unif}(\cB_a)$ sampled.
However, as discussed in Sec.~\ref{sec:autobackswap}, requiring access to spurious attributes during training is undesirable.
Therefore, we also consider sampling based on the target label $y \sim \text{Unif}(\cY)$, selecting backgrounds $\cB_y$, and drawing $\Bx_{\text{bg}} \sim \text{Unif}(\cB_y)$.
Finally, we evaluate a fully agnostic strategy that samples a background uniformly from all possible backgrounds, $\Bx_{\text{bg}} \sim \text{Unif}(\cB)$.

We assess these strategies on the Waterbirds dataset, with results reported in Tab.~\ref{tab:sampling_strategy}.
In the setting with minority samples, sampling conditioned on the spurious attribute yields the highest WGA, while target-conditioned sampling results in slightly lower Acc.
In the setting without minority samples, uniformly sampling from all backgrounds leads to reduced WGA.
Overall, performance differences across strategies are modest, though sampling conditioned on the spurious attribute provides a small but consistent advantage when available.

\section{Conclusion} \label{sec:conclusion}

In this work, we introduced \methodlong{} (\method{}), a scalable data augmentation framework designed to reduce classifier reliance on spurious background correlations. 
By disentangling foreground and background using a lightweight detector, infilling missing background regions, and recomposing novel training samples, \method{} enables the training of background-invariant classifiers without requiring access to training samples that explicitly break the spurious correlation. 
Extensive experiments across multiple benchmarks demonstrate that \method{} consistently improves worst-group accuracy and overall robustness, outperforming prior methods especially in challenging settings where minority or counterfactual groups are absent during training.

Despite these strong results, we acknowledge several limitations of our approach. 
First, \method{} requires an auxiliary dataset with patch-level foreground masks to train the detector, incurring additional labeling cost, although we show that only a few hundred samples are sufficient. 
Second, the recomposition process does not explicitly account for semantic or geometric consistency between foreground and background, which may lead to unrealistic placements in some cases.
This could be alleviated in future work by synthesizing an image from a given foreground and background by means of generative models.
Third, while Sec.~\ref{sec:ablation1} demonstrates that simple infilling strategies already yield competitive performance, the availability of a suitable generator model for background infilling in highly specialized domains may still pose challenges.

Overall, \method{} provides an effective and practical solution for mitigating spurious background reliance in visual recognition systems, offering a promising direction for improving robustness and generalization in systems deployed in real-world settings.

\vspace{-0.1cm}
\section*{Acknowledgements}
\vspace{-0.2cm}

We thank Wei Lin and Sepp Hochreiter for valuable feedback in early stages of this project.

\bibliography{literature.bib}
\bibliographystyle{plainnat}

\appendix

\section{Broader Impact} \label{sec:apx:impact}

This work addresses the problem of spurious correlations in image classifiers, with direct relevance to high-stakes deployment contexts. 
We discuss both potential positive and negative societal implications below.

\paragraph{Positive impacts.}
Deep neural networks deployed in consequential domains, medical imaging, autonomous driving, and facial recognition, among others, are known to exploit spurious background cues rather than causally relevant features, with documented harms to underrepresented groups \citep{Buolamwini:18, Oakden-Rayner:20}.
\method{} can potentially reduce the risk that classifiers fail precisely on the subpopulations that are already underrepresented in training data.
Crucially, our method requires only a small auxiliary annotation effort (a few hundred samples) and does not assume access to group labels at training time, lowering barriers to adoption in settings where collecting balanced datasets is expensive or infeasible. 
Improved robustness to background shifts could therefore contribute meaningfully to safer, fairer deployed systems.

\paragraph{Negative impacts.}
\method{} specifically targets spurious correlations that manifest in the background of images.
It does not address spurious features that are entangled with the foreground itself, nor does it guard against other failure modes such as texture bias or dataset collection artifacts unrelated to background. 
Practitioners should therefore not treat our method as a general solution to distribution shift or as a substitute for careful dataset curation and ongoing monitoring of deployed systems, but as a specialized tool the particular problem of spurious backgrounds.

Additionally, the recomposition step can produce training images in which foreground objects appear in semantically implausible contexts. 
While our experiments suggest this does not harm average accuracy, it is conceivable that, in safety-critical domains, training on unrealistic composites could introduce unexpected failure modes not captured by our benchmarks. 
We encourage practitioners to validate augmented data quality before deployment.

Finally, as with any method that improves classifier robustness, there is a risk of inducing overconfidence in its predictions: a model that is more robust to background variation may be incorrectly assumed to be robust along other axes of variation as well. 
We caution against generalizing robustness claims beyond the specific spurious background correlations \method{} is designed to address.

\section{Details on \method{}} \label{sec:apx:our_method}

\begin{figure}[b!]
    \centering
    \includegraphics[width=\linewidth]{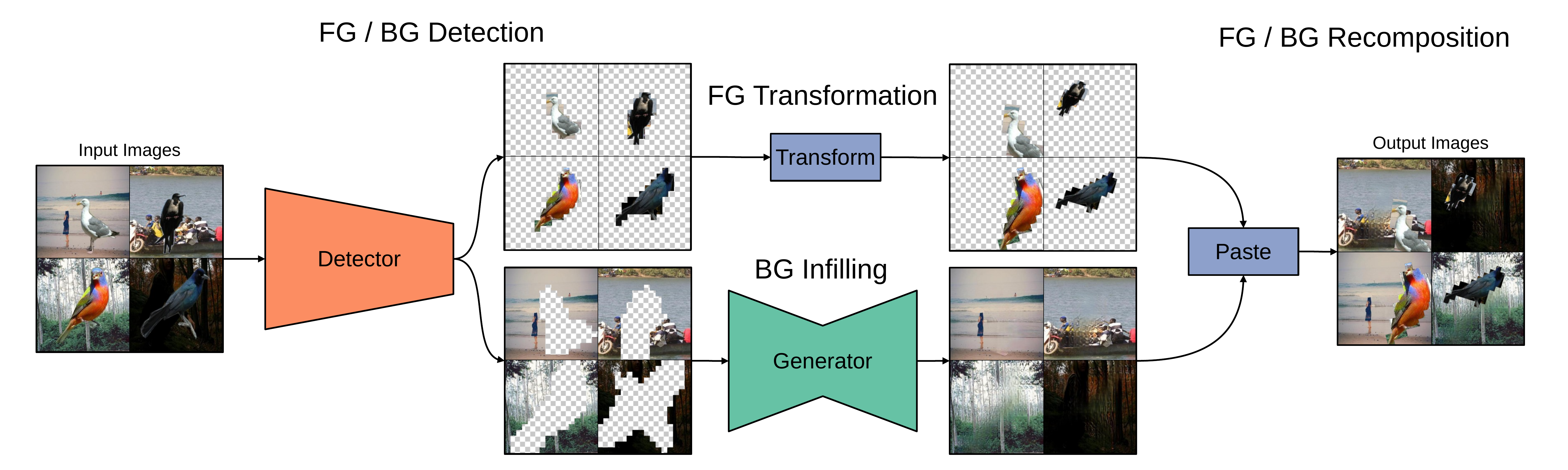}
    \caption{Illustration of applying \method{} to a dataset of four images, given an already trained detector and generator model. First, foreground and background are separated using the masks predicted by the detector. Note that two sets of masks are used, one for foreground, one for background. We would like to have little background leakage for the foreground images and little to no foreground in the background images, which is why masks are generated for two thresholds. Then, rotation, scale and translation transformations are applied on the foreground image, the background is infilled using the generator model. Finally, foreground and background are stochastically matched and recomposed to form new images, used to train the classifier.}
    \label{fig:abs_illustration}
\end{figure}

In this section, we give more implementational details of \method{}, discussing the training of the detector network, how to obtain masks for the foreground and background, and how to recompose images.
A visual overview of applying \method{} to a dataset consisting of four images is shown in Fig.~\ref{fig:abs_illustration}.

\paragraph{Detector training.}
In all experiments, we use an EfficientNet-B0 backbone and replace the final layer such that the model predicts the foreground likelihood for each output patch. 
Training is performed using binary cross-entropy on foreground/background masks. 
We optimize using SGD with momentum, where learning rate, weight decay, and momentum are selected via cross-validation using an 80:20 train/validation split of $\cD_{\text{aux}}$. 
To improve robustness, we additionally apply data augmentations (horizontal flipping, rotations, translations) and target smoothing with factor $0.1$.

\paragraph{Annotation of auxiliary labels.}
For Waterbirds, the auxiliary dataset is created manually by annotating all patches (patch resolution $16 \times 16$) that contain foreground information, as determined by the class label, resulting in 287 labeled images. 
For Spawrious, Spurious Vehicles, and for the ablation studies, we use the more scalable alternative of obtaining down-scaled masks from SAM3. 
Concretely SAM3 is prompted with the respective foreground object class to generate pixel-level segmentation masks for training images. 
These masks are not used directly during training of our method, but instead serve as proxy annotations for constructing the auxiliary dataset used to train the detector. 
We then select 400 samples and convert the pixel-level masks into coarse $16 \times 16$ patch-level labels by pooling mask values within each patch. 
These patch-wise annotations are later used to train the lightweight detector, analogous to the manually labeled Waterbirds setup.

\paragraph{Thresholds for foreground and background masks.}
Naively, one would predict a single mask separating foreground and background with the detector model.
However, we found that using two distinct masks yields better performance of the overall method in practice.
That is, because for the extracted foreground, we want as little background leakage as possible, such that the correlation between $y$ and $a$ can be broken through recomposition.
For the remaining background, we would like to certainly not contain any remaining foreground patches, lest the generator infills with foreground information again instead of generating a full background.
In our experiments, we manually tuned the thresholds used for those two masks through manually inspecting the resulting foregrounds and backgrounds on random samples from the training set.
However, this procedure can also be automated by selecting thresholds according to desired precision and recall targets for patch-level predictions on the validation split of the auxiliary dataset. 
In particular, we evaluated a simple strategy on Waterbirds that fixes target positive and negative rates on the validation set, yielding thresholds close to those obtained via manual inspection and leading to virtually identical downstream performance.

\paragraph{Recomposition.}
Finally, the obtained foregrounds and backgrounds are recomposed stochastically to form new augmented samples.
Here we apply transformations on the foreground image, rotation, rescaling and translation.

\section{Details on Baseline Methods} \label{sec:apx:baselines}

\subsection{Trained Baselines}

\paragraph{ERM.}
Empirical Risk Minimization corresponds to standard supervised training of a classifier using cross-entropy loss on the original training data, without any explicit consideration of spurious attributes or group structure.

\paragraph{ERM + Heavy Aug.}
In addition to standard data augmentation (random resized crop and horizontal flipping), we apply mixup \citep{Zhang:18}, CutMix \citep{Yun:19}, and ColorJitter. 
This baseline evaluates whether stronger generic augmentation alone can mitigate spurious correlations.

\paragraph{Chang et al.}
We implement the method of \citet{Chang:21}, which generates factual and counterfactual samples using ground-truth or estimated foreground masks and jointly optimizes three losses: standard classification loss, a loss encouraging the correct class prediction on factual samples, and a loss penalizing the correct class prediction on counterfactual samples. 
For Waterbirds, we use the provided ground-truth segmentations. 
For Spawrious, segmentation masks are obtained using SAM3, prompting it to segment ``dog''.
The original implementation uses the rather dated CAGAN model \citep{Yu:18} for infilling of the counterfactuals.
We switch to the AOT-GAN model used in \method{}, to rule out differences in the infilling quality as reason for performance differences.
Furthermore, AOT-GAN was developed as an improved successor model to CAGAN, leading to improved infilling quality in their experiments (see their Table 1).

\paragraph{DFR.}
Deep Feature Reweighting first trains a backbone model using ERM. 
In a second stage, only the final classification layer is retrained using a group-balanced reweighting dataset, which requires access to group labels.

\paragraph{AFR.}
Automatic Feature Reweighting also follows a two-stage training scheme. 
After ERM training, the final layer is retrained with samples weighted according to the prediction errors of the base model, without requiring explicit group labels.
As an approximation to DFR with reduced assumptions, it is expected to at best perform as good as DFR.
Importantly, the first stage should be ``trained until convergence'' (see their Algorithm 1).
The authors further note that in their experiments, the first stage achieves near-zero training loss and might overfit and not predict well at all on minority samples.
When experimenting with AFR, we also found that long training in the first stage was critical and in the end refrained from using early stopping as for all other methods to fairly evaluate the performance of this method.

\paragraph{Group DRO.}
Group Distributionally Robust Optimization minimizes the worst-case loss across predefined groups. 
This method requires access to spurious attribute labels during training and explicitly optimizes for worst-group performance.
In particular, we implement the online optimization algorithm (Algorithm 1 in \citet{Sagawa:20}).

\paragraph{CORAL.}
Correlation Alignment aligns second-order statistics of feature representations across domains defined by spurious attributes through an additional regularization term, encouraging background-invariant representations.
Originally developed for domain generalization, it can be readily applied to mitigate spurious correlations, but spurious attribute labels are required during training.

\subsection{Zero-Shot Baselines}

\paragraph{SAM3.}
We prompt SAM3 with the target class names and obtain segmentation masks. 
Predictions are derived by selecting the class whose predicted segmentation mask covers the largest area.

\paragraph{CLIP.}
We compute cosine similarities between the image embedding and text embeddings of class names, and predict the class with the highest similarity.
In particular, we use the Open-CLIP ViT-bigG-14 model \citep{Ilharco:21}, trained on the LAION-2B \citep{Schuhmann:22} s39b b160k.

\paragraph{QWEN3-VL-Thinking.}
We prompt the model to explicitly ignore background cues and reason about the object of interest before producing a class prediction. 
We found this prompting strategy crucial for achieving competitive performance.
The full prompts for Waterbirds and Spawrious are as follows:

\vspace{0.3cm}
\begin{tcolorbox}[
  title=QWEN3-VL-Thinking Prompt (Waterbirds),
  colback=gray!25,
  colframe=black!50,
  boxrule=0.5pt,
  arc=3pt,
  left=6pt,
  right=6pt,
  top=6pt,
  bottom=6pt
]
Does this picture show a landbird or a waterbird species? 
Ignore the background for your answer.
Provide as answer a single word from this list: landbird, waterbird.

Answer: 
\end{tcolorbox}
\vspace{0.3cm}
\begin{tcolorbox}[
  title=QWEN3-VL-Thinking Prompt (Spawrious),
  colback=gray!25,
  colframe=black!50,
  boxrule=0.5pt,
  arc=3pt,
  left=6pt,
  right=6pt,
  top=6pt,
  bottom=6pt
]
What type of dog breed is shown in this picture?
Ignore the background for your answer.
Provide as answer a single word from this list: bulldog, corgi, dachshund, labrador.

Answer: 
\end{tcolorbox}
\vspace{0.3cm}

\section{Details on Datasets} \label{sec:apx:datasets}

In this section, we provide additional details on the datasets used in our experiments, including additional details on experimental conditions.

\subsection{Waterbirds}

\begin{figure}
    \centering
    \includegraphics[width=0.4\linewidth]{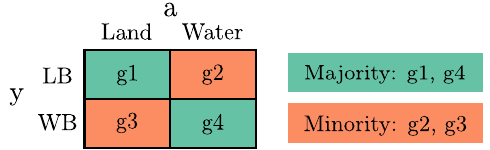}
    \caption{The Waterbirds dataset has four groups that split up into \textcolor{C0}{minority} (landbirds on water, waterbirds on land) and \textcolor{C1}{majority} (landbirds on land, waterbirds on water) groups.}
    \label{fig:waterbirds}
\end{figure}

\begin{wrapfigure}{r}{0.28\textwidth}
  \centering
  \vspace{-0.4cm}
  \includegraphics[width=0.13\textwidth, trim=0cm 13cm 39.5cm 0.8cm, clip]{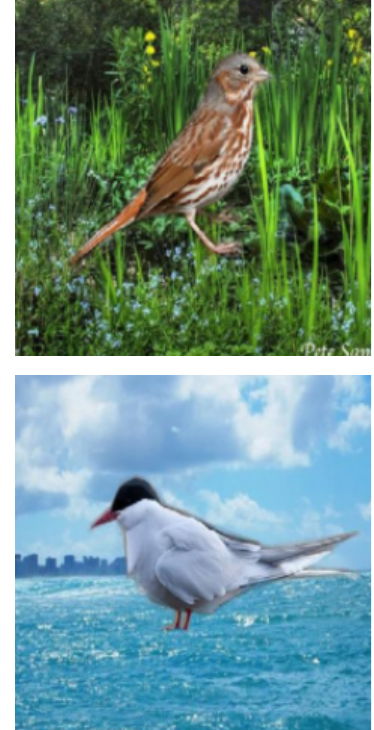}
  \includegraphics[width=0.13\textwidth, trim=0cm 0.3cm 39.5cm 13.5cm, clip]{Figures/illustrations/infilling_ablations.pdf}
  \caption{Waterbirds example images.}
  \label{fig:waterbirds_example_images}
\end{wrapfigure}

The Waterbirds dataset \citep{Sagawa:20} (Licenced under MIT) is a synthetic benchmark designed to study spurious correlations between foreground objects and background scenes. 
It consists of bird images pasted onto background images depicting either water or land, see Fig.~\ref{fig:waterbirds_example_images} for examples.
The class labels are $\cY$~=~\{waterbird, landbird\} and the spurious attribute is $\cA$~=~\{water background, land background\}, leading to four different groups.

The dataset features ground-truth segmentation maps (on the pixel level) for the birds.
Those are used for the method by \citet{Chang:21} in the main experiments and to construct certain conditions for ablations on \method{}.
\method{} does not use the ground-truth segmentation maps in the main experiments, but only relies on 287 hand-labeled patch-wise masks to train the detector for foreground / background disentanglement on the full training dataset.
Hand-labeling was done by a single annotator through a dedicated software tool written for this work, in order to optimize the workflow and save time.
Overall, it took the annotator a bit more than four hours to label all 287 images.

\newpage
\subsection{Spawrious}

\begin{wrapfigure}{r}{0.28\textwidth}
  \centering
  \vspace{-0.4cm}
  \includegraphics[width=0.26\textwidth]{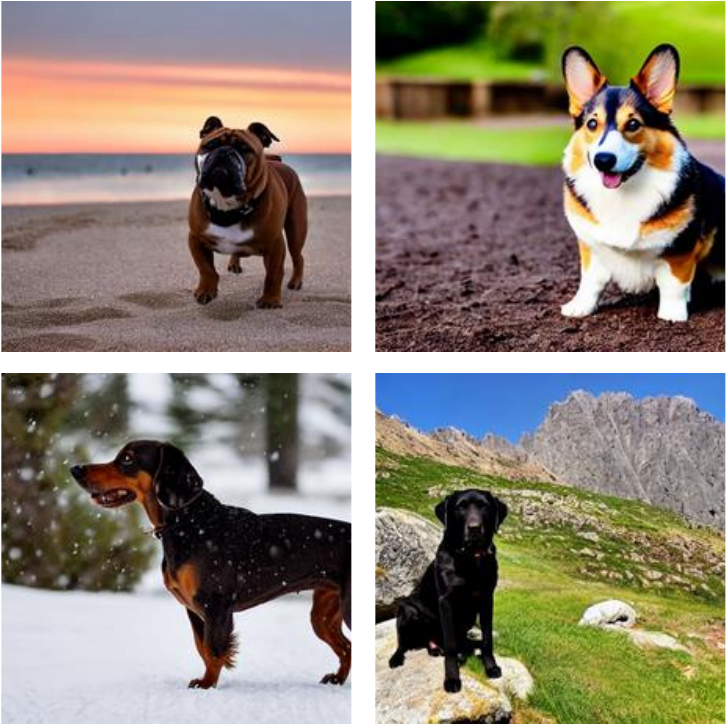}
  \caption{Spawrious example images.}
  \label{fig:spawrious_example_images}
  \vspace{-0.2cm}
\end{wrapfigure}

The Spawrious dataset \citep{Lynch:23} (Licenced under CC0 1.0 Universal) is a large-scale synthetic benchmark created using a text-to-image diffusion model. 
The aim is to classify differnet dog breeds, thus the target classes are $\cY$~=~\{bulldog, dachshund, labrador, corgi\}.
They are generated in different environments, thus the spurious background attributes are $\cA$~=~\{desert, jungle, dirt, snow, beach, mountain\}, yielding 24 possible target–background group combinations. 

\begin{figure}
    \centering
    \hfill
    \begin{subfigure}{0.3\textwidth}
    \includegraphics[width=\linewidth]{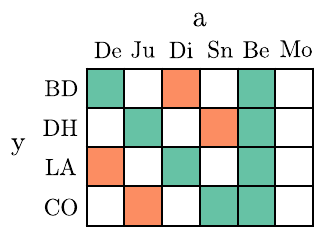}
    \caption{o2o-easy}
    \end{subfigure}
    \hfill
    \begin{subfigure}{0.3\textwidth}
    \includegraphics[width=\linewidth]{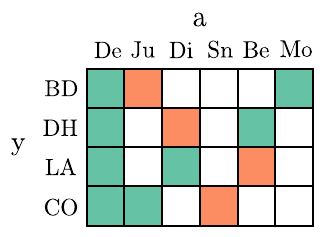}
    \caption{o2o-medium}
    \end{subfigure}
    \hfill
    \begin{subfigure}{0.3\textwidth}
    \includegraphics[width=\linewidth]{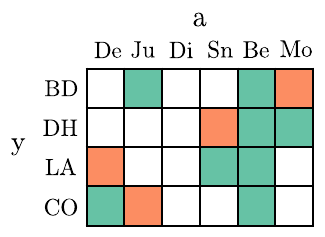}
    \caption{o2o-hard}
    \end{subfigure}
    \hfill\\
    \hfill
    \begin{subfigure}{0.3\textwidth}
    \includegraphics[width=\linewidth]{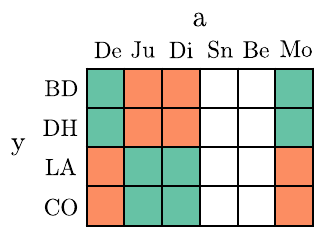}
    \caption{m2m-easy}
    \end{subfigure}
    \hfill
    \begin{subfigure}{0.3\textwidth}
    \includegraphics[width=\linewidth]{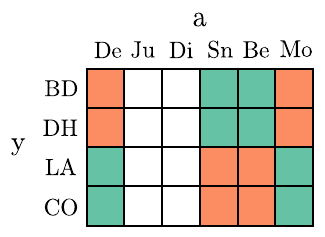}
    \caption{m2m-medium}
    \end{subfigure}
    \hfill
    \begin{subfigure}{0.3\textwidth}
    \includegraphics[width=\linewidth]{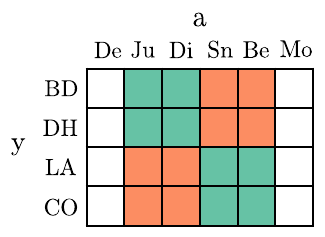}
    \caption{m2m-hard}
    \end{subfigure}
    \hfill
    \caption{Detailed configurations of Spawrious datasets. \textbf{One-to-One:} \textcolor{C0}{Train set} uses a dominant and a shared background (9:1), while \textcolor{C1}{test set} shifts the target–background associations. \textbf{Many-to-Many:} \textcolor{C0}{Train set} assigns two backgrounds per class at equal ratio, and \textcolor{C1}{test set} swaps these associations. There is a disagreement between the naming convention of the many-to-many difficulty levels in the paper and the official implementation. We chose to follow the implementation.}
    \label{fig:spawrious}
\end{figure}

The benchmark defines two experimental settings: one-to-one (o2o) and many-to-many (m2m), each with three difficulty levels (easy, medium, hard), which vary in how the target–background associations differ between training and testing.
The individual configurations are shown in Fig.~\ref{fig:spawrious}.
Note that for the many-to-many settings, there is a discrepancy between the description in the paper and the official implementation by \citet{Lynch:23}, regarding which groups belong to which difficulty level.
We chose to follow the official implementation.

Segmentation masks required for the \citet{Chang:21} baseline are obtained using SAM3, prompting it with ``dog''.
For very few images (less than 0.1\% of the dataset), it fails to segment a dog.
We find this is generally if the dog is shown very tiny in the background and hard to make out as dog definitively when not knowing that the image should contain a dog.
We excluded those images from the training set, due to their very limited number.
For \method{}, we obtained patch annotations with a resolution of 16 on 400 samples.
We chose more samples, as the Spawrious dataset is more complex than the Waterbirds dataset.
Moreover, the training set contains different groups in each of the six experimental conditions (o2o-easy, o2o-medium, ...).
Therefore, we automatically generated the patch-level annotations based on the pixel-level segmentation masks obtained by SAM3.
To be clear, if segmentation masks are already available for the whole training dataset, there is no point to train a detector in practice, one can simply use the segmentation masks. 
We do this to emulate not having access to segmentation masks and obtaining the patch-level masks through manual annotation.

\newpage
\subsection{Spurious Vehicles} \label{sec:apx:spurious_vehicles}

\begin{wrapfigure}{r}{0.28\textwidth}
  \centering
  \vspace{-0.4cm}
  \includegraphics[width=0.26\textwidth]{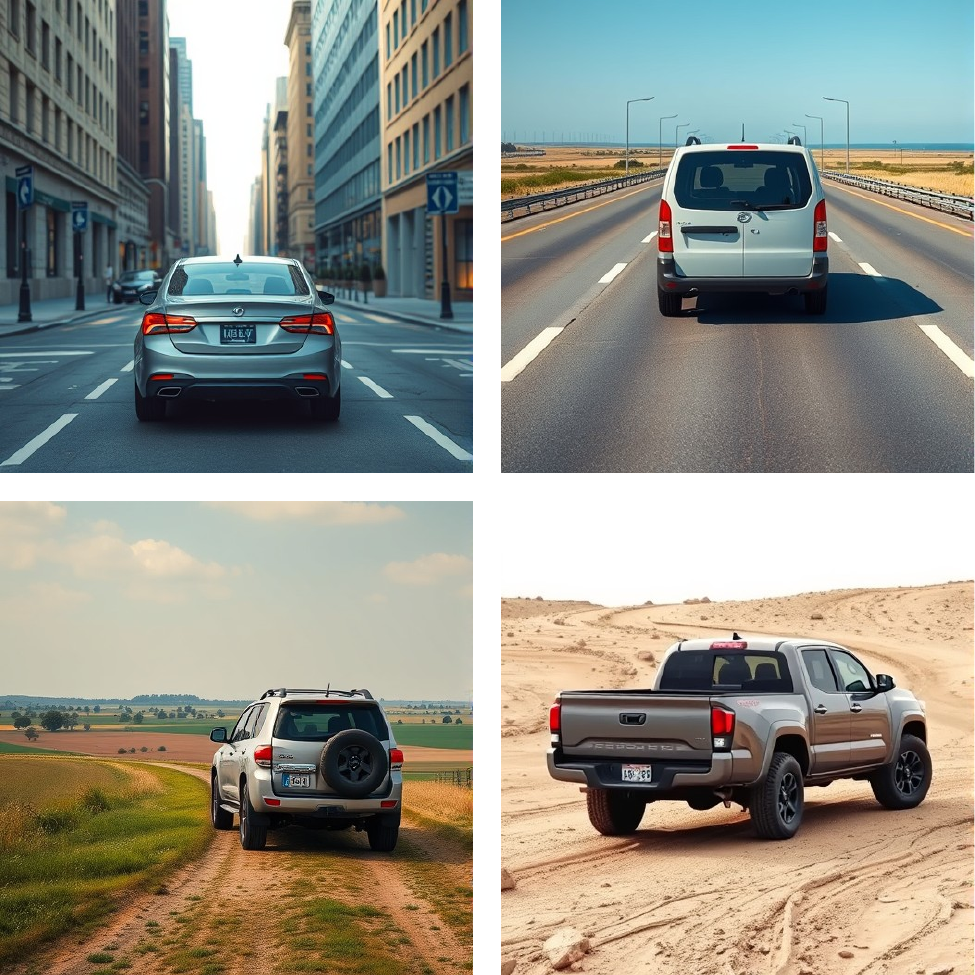}
  \caption{Spurious Vehicles example images.}
  \label{fig:vehicles_example_images}
  \vspace{-0.2cm}
\end{wrapfigure}

The \textit{Spurious Vehicles} dataset is a synthetic benchmark introduced in this work, generated using the FLUX.1-schnell model by \citet{BlackForestLabs:25} (Licensed under Apache-2.0). The aim is to classify different vehicle types, thus the target classes are $\cY$~=~\{\text{sedan}, \text{minivan}, \text{SUV}, \text{pickup truck}\}. Images are generated in different environments, yielding the spurious context attributes $\cA$~=~\{\text{urban}, \text{highway}, \text{rural}, \text{off-road}\}, for a total of 16 possible target--context group combinations.

\begin{figure}
    \centering
    \includegraphics[width=0.35\textwidth]{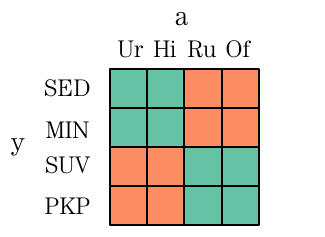}
    \caption{Detailed configuration of the Spurious Vehicles dataset (many-to-many setting). The \textcolor{C0}{training set} assigns two background contexts to each class in equal proportion, while the \textcolor{C1}{test set} swaps these associations.}
    \label{fig:spurious_vehicles_configurations}
\end{figure}

We consider the many-to-many setting, analogous to Spawrious \citep{Lynch:23}. Each target class is associated with two contexts during training, where pairs of classes share the same contexts. At test time, these assignments are swapped while preserving the shared structure. The exact configuration is shown in Fig.~\ref{fig:spurious_vehicles_configurations}. We generate 256 images per group, resulting in 4096 images overall, with half belonging to the training/validation split and half to the test split.

Images are generated at a resolution of $512 \times 512$ using a quantized FLUX.1-schnell checkpoint (Q8\_0 format) for efficient inference. We use guidance scale $0.0$, four denoising steps, and a maximum sequence length of 256. Prompts specify both the vehicle class and the desired context while enforcing consistent framing and viewing perspectives. This controlled generation process allows us to vary contextual cues while preserving the semantic target label.

\section{Hyperparameter Tuning} \label{sec:apx:hyperparameters}

We manually tuned \method{} and baselines using both WGA and Acc on the validation dataset. 
On the Waterbirds benchmark, we tuned in the with minority setting and used the same hyperparameters for the setting without minority groups.
On the Spawrious benchmark, we used the o2o-easy setting and used the same hyperparameters for all other settings.
Particularly, we tuned learning rate, batch size and weight decay for each method, in addition to important method specific parameters (e.g. regularization weight for CORAL or group-weight step size for GroupDRO).
Many other hyperparameters, such as the optimizer (SGD), its momentum (0.9), the early stopping strategy, or the strength of target smoothing are derived from prior work \citep[e.g.][]{Sagawa:20, Kirichenko:23, Qiu:23} and shared across all methods. 
We investigated how well they are tuned on the ERM baseline and found no significantly better configuration.
Where available, we started from the hyperparameters originally proposed, or from hyperparameters used for the particular method on the particular dataset in prior work.
Given the similar dataset structure, we follow a comparable hyperparameter configuration for the Spurious Vehicles benchmark as for Spawrious.
All methods have been tuned on the same seed, with a comparable number of configurations

\section{Effect of Model Architecture} \label{sec:apx:waterbirds_vit}

In addition to the ResNet50 architecture \citep{He:16} used for the results in the main paper, we further consider the vision transformer (ViT) architecture \citep{Dosovitskiy:21} to assess the robustness of our results to changing architectures.
In particular, we consider the same experimental setup as in Sec.~\ref{sec:waterbirds}.

We did not re-tune hyperparameters compared to the Waterbirds experiments in the main paper, but used a general recipe following standard practice to port the configurations of all methods from ResNet-50 to ViT.
In particular, we changed the optimizer from SGD to AdamW \citep{Loshchilov:19}, decreasing the learning rate by 1e-2 and increasing weight decay by 1e1.
Furthermore, we add a learning rate scheduler of 5 epochs of linear warmup (from 10\% of start learning rate), followed by cosine decay to 1\% of the start learning rate.
Note that for \method{} only the classifier is changed to a ViT, the auxiliary detector model is still a ResNet-50, as we care about the performance of the classifier when trained on the augmented samples, not how to optimally train a foreground / background detector with ViTs which may require different training strategies.

\paragraph{Results.}
\method{} attains slightly higher Acc with ViT than with ResNet, but slightly lower WGA.
All baseline methods substantially decrease in performance when changing to ViTs, further increasing the advantage of \method{} over the baseline methods.
\method{} improves \textbf{44.3\%} in WGA and \textbf{19.2\%} in Acc over the second best method in the setting without minority samples.

\setlength{\tabcolsep}{2.0pt}
\renewcommand{\arraystretch}{1.1}
\begin{table}
\centering
\caption{Extended results for Waterbirds experiments featuring both results with ResNet50 (same as Tab.~\ref{tab:waterbirds_main}) and ViT as base models. We report the \textit{WGA} and the \textit{Acc} over the four groups. Best result \textbf{bold}, second best \underline{underlined}. Statistics computed over five independent runs.\vspace{0.1cm}}
\label{tab:waterbirds_vit}
\small
\begin{tabular}{cccccccccccccccc}
\toprule
\multirow{3}{*}{\textbf{Method}} && \multicolumn{5}{c}{\textbf{ResNet50}} &&& \multicolumn{5}{c}{\textbf{ViT}} \\ 
&& \multicolumn{2}{c}{\textbf{w minority}} && \multicolumn{2}{c}{\textbf{w/o minority}} &&& \multicolumn{2}{c}{\textbf{w minority}} && \multicolumn{2}{c}{\textbf{w/o minority}}  \\ \cline{3-4}\cline{6-7}\cline{10-11}\cline{13-14}
\textbf{}                         &&  \textit{WGA}           & \textit{Acc}           &&  \textit{WGA}            & \textit{Acc}  &&&  \textit{WGA}           & \textit{Acc}           &&  \textit{WGA}            & \textit{Acc}      \\ \midrule
ERM                               &&           $74.9_{(3.3)}$           &           $93.6_{(0.4)}$      &&          $33.4_{(4.4)}$           &      $68.4_{(1.9)}$    &&&           $63.1_{(10.4)}$           &           $90.1_{(0.6)}$      &&          $21.4_{(1.3)}$           &      $62.5_{(1.8)}$             \\
+ Heavy Aug.                   &&       $74.3_{(2.3)}$              &           $87.4_{(1.0)}$         &&            $15.9_{(2.7)}$          &          $61.5_{(1.2)}$      &&&    $66.0_{(5.8)}$    &   $90.8_{(0.7)}$    &&   $19.4_{(3.4)}$   &         $60.2_{(1.5)}$       \\
DFR                               &&           $\underline{90.7}_{(0.8)}$          &         $94.3_{(0.1)}$           &&         \textcolor{black}{$15.1_{(1.9)}$}             &           \textcolor{black}{$60.1_{(0.8)}$}        &&&     $\underline{87.7}_{(1.1)}$   &   $92.1_{(0.6)}$    &&   $20.9_{(4.9)}$    &     $61.6_{(1.9)}$       \\
AFR                               &&             $78.6_{(3.7)}$        &              $88.5_{(0.9)}$      &&             \textcolor{black}{$22.2_{(3.3)}$}         &           \textcolor{black}{$64.2_{(0.8)}$}         &&&    $59.9_{(3.1)}$    &   $79.3_{(1.5)}$    &&    $19.9_{(2.3)}$  &     $60.0_{(1.3)}$       \\
Group DRO                         &&            $82.2_{(1.6)}$          &         $94.8_{(0.2)}$           &&            $27.6_{(3.8)}$          &           $65.3_{(0.6)}$         &&&    $80.7_{(5.2)}$    &   $93.3_{(0.7)}$    &&   $20.4_{(4.8)}$   &     $61.4_{(2.0)}$
       \\
CORAL                             &&             $78.4_{(4.2)}$        &          $93.6_{(0.3)}$          &&           $22.0_{(5.5)}$           &            $65.0_{(2.3)}$        &&&    $64.1_{(8.0)}$    &    $90.7_{(0.5)}$   &&    $20.8_{(2.8)}$  &       $60.7_{(1.5)}$     \\ 
Chang et al. &&          $85.7_{(2.3)}$           &           $\boldsymbol{95.7}_{(0.3)}$         &&           $79.1_{(3.2)}$           &           $91.4_{(1.8)}$         &&&    $78.5_{(3.6)}$    &     $\underline{93.7}_{(0.5)}$  &&   $\underline{45.5}_{(8.0)}$   &      $\underline{76.4}_{(4.3)}$      \\
\rowcolor{gray!15} \textbf{Ours}        &&           $\boldsymbol{93.0}_{(0.9)}$          &         $\underline{95.0}_{(0.5)}$           &&            $\boldsymbol{92.2}_{(0.4)}$           &              $\boldsymbol{95.1}_{(0.3)}$     &&&    $\boldsymbol{90.8}_{(0.5)}$    &    $\boldsymbol{96.6}_{(0.3)}$   &&   $\boldsymbol{89.8}_{(1.1)}$   &     $\boldsymbol{95.6}_{(0.4)}$   \\ \hline
\end{tabular}
\end{table}
\renewcommand{\arraystretch}{1}

\section{Ablations}

\subsection{Quality of Auxiliary Dataset} \label{sec:apx:ablation_ds_pr}

\begin{figure*}[t]
    \centering
    \includegraphics[width=1.0\textwidth]{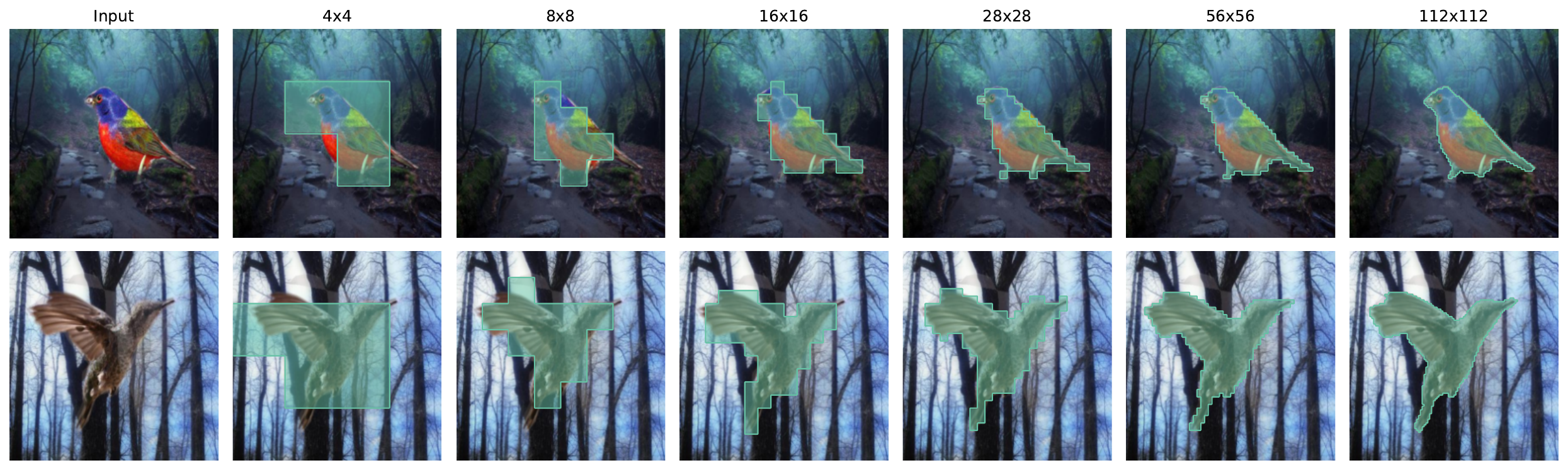}
    \caption{Visualization of mask annotations across patch resolutions. The first column shows the input image without any mask, followed by the corresponding mask overlays at different patch resolutions. The figure illustrates how decreasing the patch resolution results in coarser and more spatially aggregated masks, while increasing the patch resolution results in more spatial detail.}
    \label{fig:ablation_detector_quality_examples}
\end{figure*}

The quality of the auxiliary dataset is determined by its patch resolution and size.
Generally, the auxiliary dataset is randomly sampled from the full training dataset, followed by manual labeling in the main experiments.
Within this ablation, we use the provided ground-truth segmentation masks for Waterbirds to analyze the influence of the auxiliary datasets quality on the overall performance.
In Fig.~\ref{fig:ablation_detector_quality_examples}, we show examples of how masks look like for a certain patch resolution.
Note that a threshold of 25\% was used to assign the patches, i.e. at least 25\% of pixels must be segmented as bird.

\subsection{Background Infilling Strategy} 

We use the AOT-GAN model for background infilling for our main experiments.
Within this ablation, we investigate more simple infilling strategies that do not need a pretrained generator model:
\begin{itemize}[leftmargin=*, topsep=0pt, itemsep=4pt, partopsep=0pt, parsep=0pt]
    \item \textbf{Patch-scramble:} extract patches of the patch resolution from the remaining background and randomly insert into the empty foreground region for infilling.
    \item \textbf{Pixel-scramble:} randomly insert pixels from the remaining background into the empty foreground region for infilling.
    \item \textbf{Average:} fill the empty foreground region with the average pixel value of the remaining background.
    \item \textbf{Constant:} fill the empty foreground region with a constant value (we use 128 in uint8 pixel space, which is grey).
    \item \textbf{Noise:} following \citet{Chang:21}, we first sample uniform per color channel and add Gaussian noise on the pixel level on top of the uniform noise.
\end{itemize}

\begin{figure*}[b]
    \centering
    \includegraphics[width=1.0\textwidth]{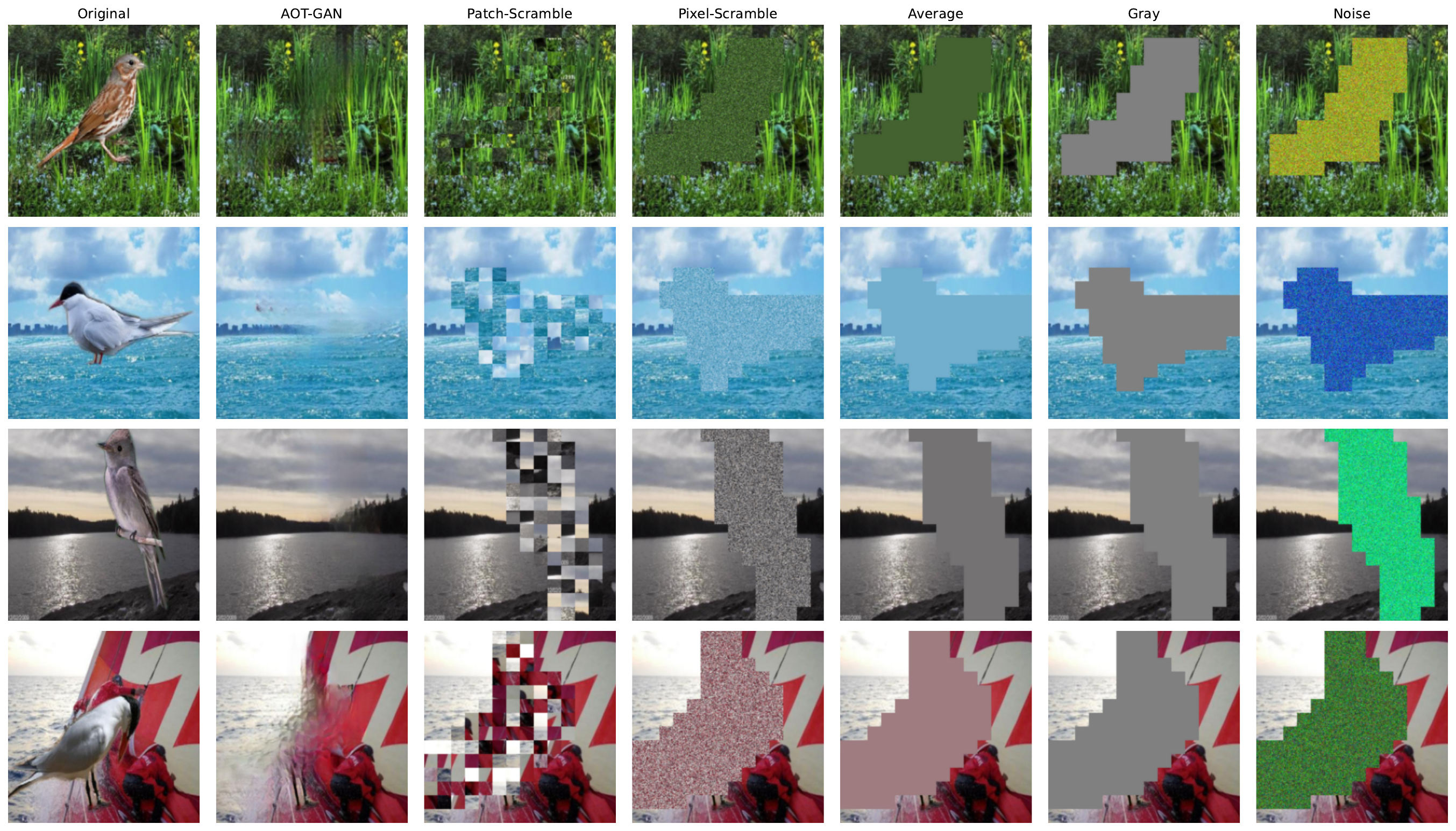}
    \caption{Qualitative comparison of background infilling strategies. The first column shows the original image, and the remaining columns correspond to different background infilling methods (AOT-GAN, Patch-Scramble, Pixel-Scramble, Average, Gray and Noise), while each row shows a different example image.}
    \label{fig:waterbirds_infill_ablation_examples}
\end{figure*}

In Fig.~\ref{fig:waterbirds_infill_ablation_examples} we show examples for the considered infilling methods.

\subsection{Biased Auxiliary Mask Labels}

In this ablation, we consider the effect of biased auxiliary mask labels.
Depending on instructions and personal biases, the patch-level mask labels could be oversampled (containing much background) or undersampled (not contain parts of the foreground).

\begin{figure*}
    \centering
    \includegraphics[width=1.0\textwidth]{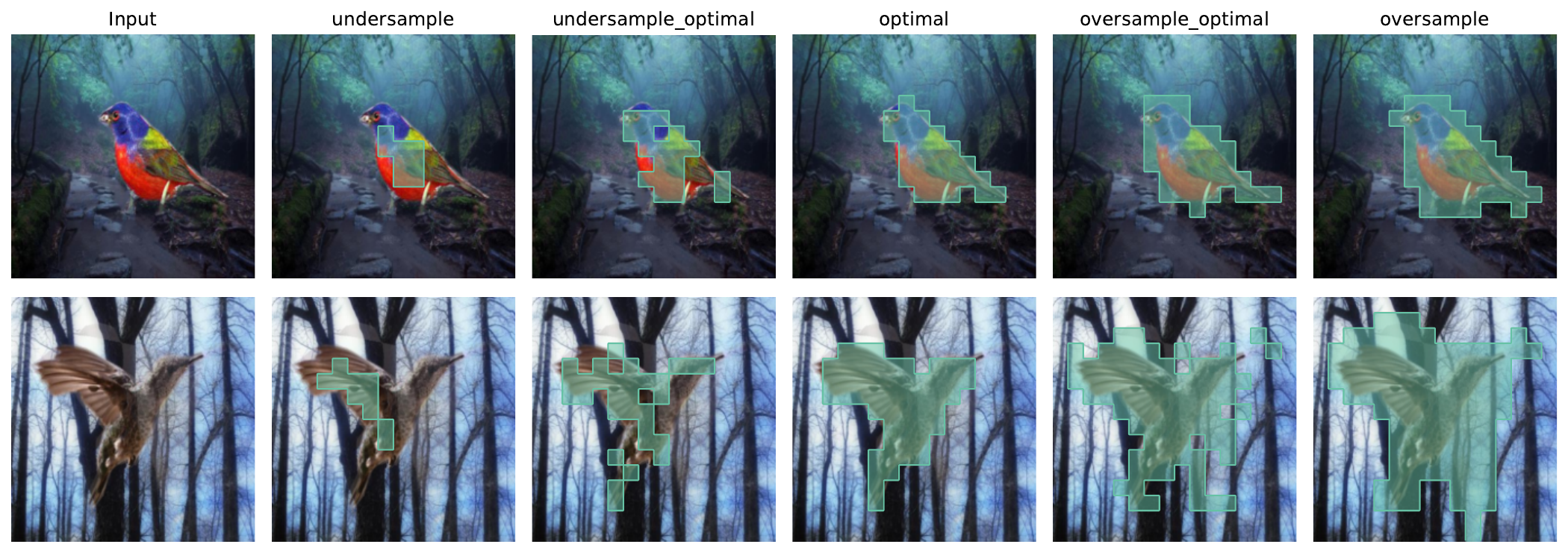}
    \caption{Visualization of labeling strategies for auxiliary mask labels. In undersampling, only patches fully containing the bird are labeled foreground. In oversampling,  a single foreground pixel suffices to label the patch as foreground. In optimal sampling, patches are labeled as foreground if at least 25\% of pixels belong to the bird. Mixtures stochastically interpolate (with p=0.5) patches between undersampling and optimal, as well as oversampling and optimal.}
    \label{fig:ablation_label_quality_examples}
\end{figure*}

Based on the ground-truth segmentation masks provided for the waterbirds dataset, we do the following patch assignments:
For oversampling, a patch is assigned as foreground if a single pixel is a foreground according to the segmentation mask.
For undersampling, a patch is assigned as foreground if every pixel is a foreground according to the segmentation mask.
Furthermore, we emulate our manual sampling outcome, by setting a threshold of 25\% of pixels containing foreground according to the segmentation mask, which we call optimal.
In addition to this basic settings, we also consider mixtures between oversampling and optimal, as well as undersampling and optimal.
For these mixtures, we randomly select each patch from either of the two base masks with equal probability.
We show examples of all settings in Fig.~\ref{fig:ablation_label_quality_examples}.

\section{Detailed Comparison to Prior Work} \label{sec:detailed_comparison}

\subsection{Chang et al. (2021)}

\citet{Chang:21} propose counterfactual and factual/invariant data augmentation based on ground-truth bounding boxes or segmentation masks. 
Their method generates augmented factual and counterfactual samples and optimizes two additional auxiliary losses on those. 
In contrast, \method{} requires only a small auxiliary dataset with coarse patch-level labels and uses it to automatically generate large-scale augmented data. 
Our approach is more straightforward, avoids additional loss terms that may be hard to balance in practice, and removes the need for dense foreground annotations on the full dataset.

\subsection{Wang et al. (2025)}

\citet{Wang:25} introduce a background-mixing strategy called BackMix based on class activation maps, where foreground regions are masked and background patches are extracted and pasted onto target images. 
This is opposite to our approach, where the foreground region is extracted and pasted onto infilled backgrounds.
Their approach focuses on open-set recognition and relies on heuristic foreground estimation. 
\method{} instead learns an explicit foreground–background disentanglement model and performs systematic recomposition, aiming for improved robustness against spurious backgrounds in closed-set classification.

\subsection{Bransby et al. (2024)}

\citet{Bransby:24} propose a background-mixing strategy also called BackMix for echocardiography, assuming access to foreground masks in a semi-supervised medical imaging setting. 
Their method samples random backgrounds for each foreground, where infilling of the remaining background can be trivially done by inserting zeros due to the data structure and applying additional weighting to those augmented samples in the cross-entropy loss. 
\method{} generalizes this idea by learning foreground masks from limited patch-level labels, enabling scalable application across diverse natural-image benchmarks without dense supervision.
It augments all training samples and does not alter the training loss.

\section{Software and Hardware} \label{sec:compute}

The code to reproduce all our experiments are available at \url{https://github.com/cesro7/spur-corr}.

We use NumPy 1.26 \citep{Harris:20}, scikit-learn 1.5 \citep{Pedregosa:11}, PyTorch 2.4 \citep{Paszke:19} and HF transformers 5.0 \citep{Wolf:20}.

The majority of experiments and hyperparameter tuning (partly) have been performed on A40 (48GB) GPUs.
Early experimentation and hyperparameter tuning (partly) been performed on RTX 4060 GPUs.
For the main experiments, including all baselines, we estimate a computational demand of around 300 GPU-hours.
Another 100 GPU-hours have been used for early experimentation and hyperparameter tuning.

\end{document}